\documentclass[runningheads]{llncs}
\usepackage{graphicx}
\usepackage{amsmath,amssymb} 
\usepackage{color}

\usepackage{multirow}
\usepackage[colorlinks]{hyperref}
\DeclareMathOperator*{\argmax}{arg\,max}
\DeclareMathOperator*{\argmin}{arg\,min}

\begin{document}
\title{A Deeper Look at 3D Shape Classifiers} 

\titlerunning{A Deeper Look at 3D Shape Classifiers}
%
\author{Jong-Chyi Su, Matheus Gadelha, Rui Wang, Subhransu Maji}
\institute{University of Massachusetts, Amherst \\ 
\email{\{jcsu,mgadelha,ruiwang,smaji\}@cs.umass.edu}}
\authorrunning{Jong-Chyi Su, Matheus Gadelha, Rui Wang, Subhransu Maji}
%
%
\maketitle              
\begin{abstract}
We investigate the role of representations and architectures for classifying 3D shapes in terms of their computational efficiency, generalization, and robustness to adversarial transformations.
By varying the number of training examples and employing cross-modal transfer learning we study the role of initialization of existing deep architectures for 3D shape classification. 
Our analysis shows that multiview methods continue to offer the best generalization even without pretraining on large labeled image datasets, and even when trained on simplified inputs such as binary silhouettes.
Furthermore, the performance of voxel-based 3D convolutional networks and point-based architectures can be improved via cross-modal transfer from image representations.
Finally, we analyze the robustness of 3D shape classifiers to adversarial transformations and present a novel approach for generating adversarial perturbations of a 3D shape for multiview classifiers using a differentiable renderer.
We find that point-based networks are more robust to point position perturbations while voxel-based and multiview networks are easily fooled with the addition of imperceptible noise to the input.
\end{abstract}
\def\point{P}
\def\matrix{P}
\def\npoints{N}
\def\nshapes{S}
\def\nbasis{B}
\def\depth{D}
\def\scale{k}
\def\encoding{\mathbf{z}}
\newcommand{\para }[1]{\medskip \noindent {\bf #1}}
\def\mrtnet{MRTNet\xspace}
\def\mrvae{MR-VAE\xspace}
\def\kdtree{kd-tree\xspace}

\section{Introduction}
\label{s:intro}
Techniques for analyzing 3D shapes are becoming increasingly important due to the vast number of sensors that are capturing 3D data, as well as numerous computer graphics applications.
In recent years a variety of deep architectures have been approached for classifying 3D shapes.
These range from multiview approaches that render a shape from a set of views and deploy image-based classifiers, to voxel-based approaches that analyze shapes represented as a 3D occupancy grid, to point-based approaches that classify shapes represented as collection of points.
However, there is relatively little work that studies the tradeoffs offered by these modalities and their associated techniques. 

This paper aims to study three of these tradeoffs, namely the ability to generalize from a few examples, computational efficiency, and robustness to adversarial transformations.
We pick a representative technique for each modality. For multiview representation we choose the Multiview CNN (MVCNN) architecture~\cite{mvcnn}; For voxel-based representation we choose the VoxNet~\cite{voxnet,modelnet} constructed using convolutions and pooling operations on a 3D grid; For point-based representation we choose the PointNet architecture~\cite{pointnet}. 
The analysis is done on the widely-used ModelNet40 shape classification benchmark~\cite{modelnet}.

Some of our analysis leads to surprising results.
For example, with deeper architectures and a modification in the rendering technique that renders with black background and better centers the object in the image the performance of a vanilla MVCNN can be improved to \textbf{95.0\%} per-instance accuracy on the benchmark, outperforming several recent approaches.
Another example is that while it is widely believed that the strong performance of MVCNN is due to the use of networks pretrained on large image datasets (e.g., ImageNet~\cite{ILSVRC15}), we find that even without such pretraining the MVCNN obtains \textbf{91.3\%} accuracy, outperforming several voxel-based and point-based counterparts that also do not rely on such pretraining.
Furthermore, the performance of MVCNN remains at \textbf{93.6\%} even when trained with binary silhouettes (instead of shaded images) of shapes, suggesting that shading offer relatively little extra information on this benchmark for MVCNN.

We then systematically analyze the generalization ability of the models.
First we analyze the accuracy of various models by varying the number of training examples per category. 
We find that the multiview approaches generalize faster obtaining near optimal performance with far fewer examples compared to the other approaches.
We then analyze the role of initialization of these networks. 
As 3D shape datasets are currently lacking in comparison to large image datasets, we employ cross-modal distillation techniques~\cite{Hinton14,Gupta_CVPR2016} to guide learning.
In particular we use representations extracted from pretrained MVCNNs to guide learning of voxel-based and point-based networks. 
Cross-modal distillation improves the performance of VoxNet and PointNet, especially when training data is limited.

Finally we analyze the robustness of these classifiers to adversarial perturbations. While generating adversarial inputs to VoxNet and PointNet is straightforward, it is not the case for multiview methods due to the rendering step. To this end we design an end-to-end differentiable MVCNN that takes an input a voxel representation and generates a set of views using a differentiable renderer.
We analyze the robustness of these networks by estimating the amount of perturbation needed to obtain a misclassification. 
We find that PointNet is more robust, while MVCNN and VoxNet are both easily fooled by adding a small amount of noise.
This is similar to the observations in prior work of adversarial inputs for image-based networks~\cite{intriguingnn2013,explainingadv2015,physicaladv2016}.
Somewhat surprisingly ImageNet pretraining reduces the robustness of the MVCNNs to adversarial perturbations.

In summary, we performed a detailed analysis of several recently proposed approaches for 3D shape classification. 
This resulted in a new state-of-the-art of \textbf{95.0\%} on the ModelNet40 benchmark.
The technical contributions include the use of cross-modal distillation for improving networks that operate on voxel-based and point-based representations and a novel approach for generating adversarial inputs for 3D shape classification for multiview approaches using a differentiable renderer.
This allows us to directly compare and generate adversarial inputs for voxel-based and view-based methods using gradient-based techniques.
The conclusion is that while PointNet architecture is less accurate, the use of orderless aggregation mechanism likely makes it more robust to adversarial perturbations compared to VoxNet and MVCNN, both of which are easily fooled.
\section{Method}
\label{sec:method}
This Section describes the protocol for evaluating the performance of various classifiers. We describe the dataset, performance metrics, and training setup in Section~\ref{sec:setup}, followed by the details of the deep classifiers we consider in Section~\ref{sec:models}, and the approach for generating adversarial examples in Section~\ref{sec:advers}.

\subsection{3D Shape Classification}
\para{Classification benchmark.}
\label{sec:setup}
All our evaluation is done on the ModelNet40 shape classification benchmark~\cite{modelnet} following the standard training and test splits provided in the dataset. 
There are 40 categories with 9483 training models and 2468 test models. 
The numbers of models are not equal across classes hence we report both the per-instance and per-class accuracy on the test set. While most of the literature report results by training on the entire training set, some earlier work, notably~\cite{mvcnn} reports results on training and evaluation on a subset consisting of 80 training and 20 test examples per category.

\begin{figure}[t]
  \begin{tabular}{c c  cccc}
      \includegraphics[width=0.16\linewidth]{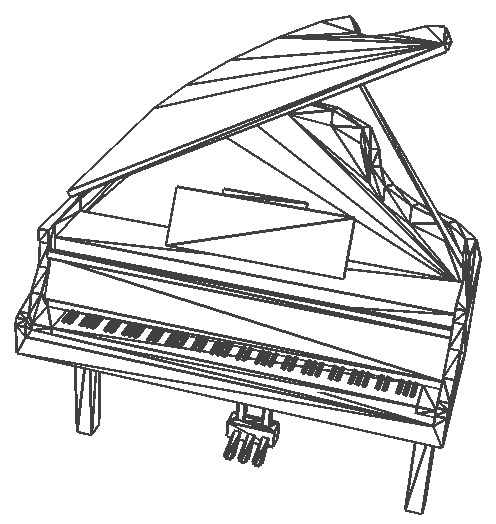}   & 
      \includegraphics[width=0.16\linewidth]{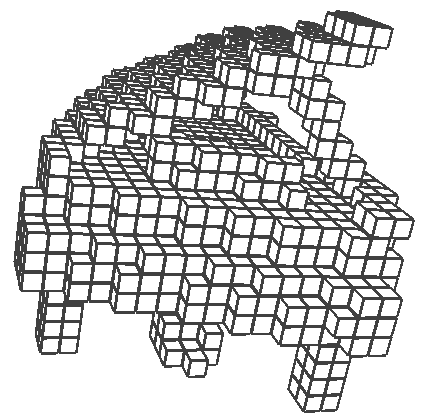}  & 
      \includegraphics[width=0.16\linewidth]{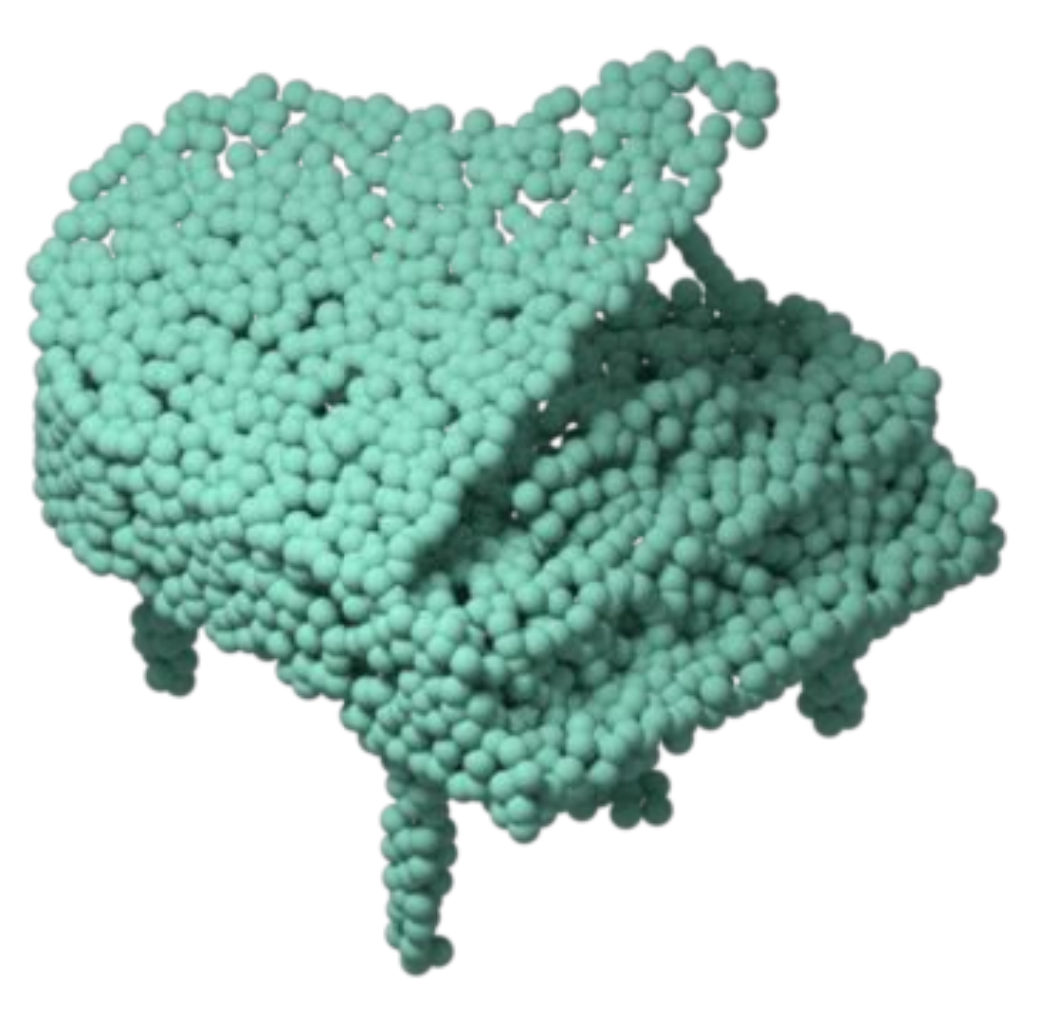} & 
      \includegraphics[width=0.16\linewidth]{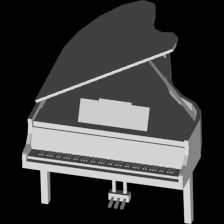} & 
      \includegraphics[width=0.16\linewidth]{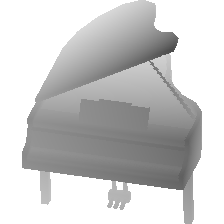} & 
      \includegraphics[width=0.16\linewidth]{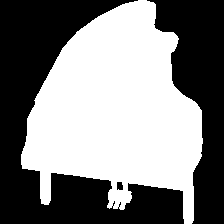} \\
	(a) Input & (b) Voxel & (c) Point cloud & (d) Phong & (e) Depth & (f) Silhouette 
  \end{tabular}
  \caption{Different representations of the input shape. From the left  (a) the shapes in the database are represented as triangle meshes, (b) the shape converted to a $30^3$ voxel grid, (c) point cloud representation with 2048 points, and (d-f) the model rendered using Phong shading, and as depth and binary silhouette images respectively.}
\label{fig:inputs}
\end{figure}

\para{Input representations.} The dataset presents each shape as a collection of triangles, hence it is important to describe the exact way in which these are converted to point clouds, voxels, and images for input to different network architectures. These inputs are visualized in Figure~\ref{fig:inputs} and described below:
\begin{itemize}
\item \textbf{Voxel representation.} To get voxel representations we follow the dataset from~\cite{qi2016volumetric} where models are discretized to a $30\times30\times30$ occupancy grid. The data is available from the author's page.
\item \textbf{Point cloud.} For point cloud representation we use the data from the PointNet approach~\cite{pointnet} where 2048 points are uniformly sampled for each model. 
\item \textbf{Image representation.} To generate multiple views of the model we use a setup similar to~\cite{mvcnn}.
Since the models are assumed to be upright oriented a set of virtual cameras are placed at 12 radially symmetric locations, i.e. every 30 degrees, facing the object center and at an elevation of 30 degrees.
Comparing to~\cite{mvcnn}, we render the images with black background and set the field-of-view of the camera such that the object is bounded by image canvas and rendered as an image of size $224\times 224$. A similar scheme was used to generate views for semantic segmentation of shapes in the Shape PFCN approach~\cite{kalogerakis20173d}. This had a non-negligible impact on the performance of the downstream models as discussed in Section~\ref{sec:few_examples}. Given the setup we considered three different ways to render the models described below:
\begin{enumerate}
\item Phong shading, where images are rendered with the Phong reflection model~\cite{phong1975illumination} using Blender software~\cite{blender}. The light and material setup is similar to the approach in~\cite{mvcnn}. 
\item Depth rendering, where only the depth value is recorded.
\item Silhouette rendering, where images are rendered as binary images for pixels corresponding to foreground.
\end{enumerate}
\end{itemize}

\para{Data augmentation.} Models in the dataset are upright oriented, but not consistently oriented along the axis, i.e., models could be rotated arbitrarily along the upright direction. 
Models that rely on voxel or point cloud input often benefit from rotation augmentation along the upright axis during training and testing.
Similar to the multiview setting we consider models rotated by 30 degree increments as additional data during training, and optionally aggregating votes across these instances at test time.

\subsection{Classification Architectures} 
\label{sec:models}
We consider the following deep architectures for shape classification.
\subsubsection{Multiview CNN (MVCNN)}
The MVCNN architecture~\cite{mvcnn} uses rendered images of the model from different views as input. Each image is fed into a CNN with shared weights.
A max-pooling layer across different views is used to perform an orderless aggregation of the individual representations followed by several non-linear layers for classification. While the original paper~\cite{mvcnn} used the VGG-M network~\cite{Simonyan14c} we also report results using:
\begin{itemize}
\item The VGG-11 network, which is the model with configuration A from~\cite{Simonyan14c}. The view-pooling layer is added before the first \texttt{fc} layer.
\item Variants of residual networks proposed in~\cite{he2016deep} such as ResNet18, ResNet34, and ResNet50. The view-pooling layer is added before the final \texttt{fc} layer.
\end{itemize}


\subsubsection{Voxel network (VoxNet)}
The VoxNet was first proposed in several early works~\cite{voxnet,modelnet} that uses convolution and pooling layers defined on 3D voxel grids.
The early VoxNet models~\cite{voxnet} used two 3D convolutional layers and 2 fully-connected layers. In our initial experiments we found the capacity of this network is limited. We also experimented with the deeper VoxNet architecture proposed in~\cite{pointnet} which has five blocks of (\texttt{conv3d-batchnorm-LeakyReLU}) and includes batch normalization~\cite{ioffe2015batch}. All \texttt{conv3d} layers have kernel size 5, stride 1 and channel size 32. The \texttt{LeakyReLU} has slope $0.1$. Two fully-connected layers (\texttt{fc-batchnorm-ReLU-fc}) are added on top to obtain class predictions. 

\subsubsection{VoxMVCNN} \label{sec:VoxMVCNN} 
\begin{figure}[t]
\centering
\includegraphics[width=0.80\linewidth]{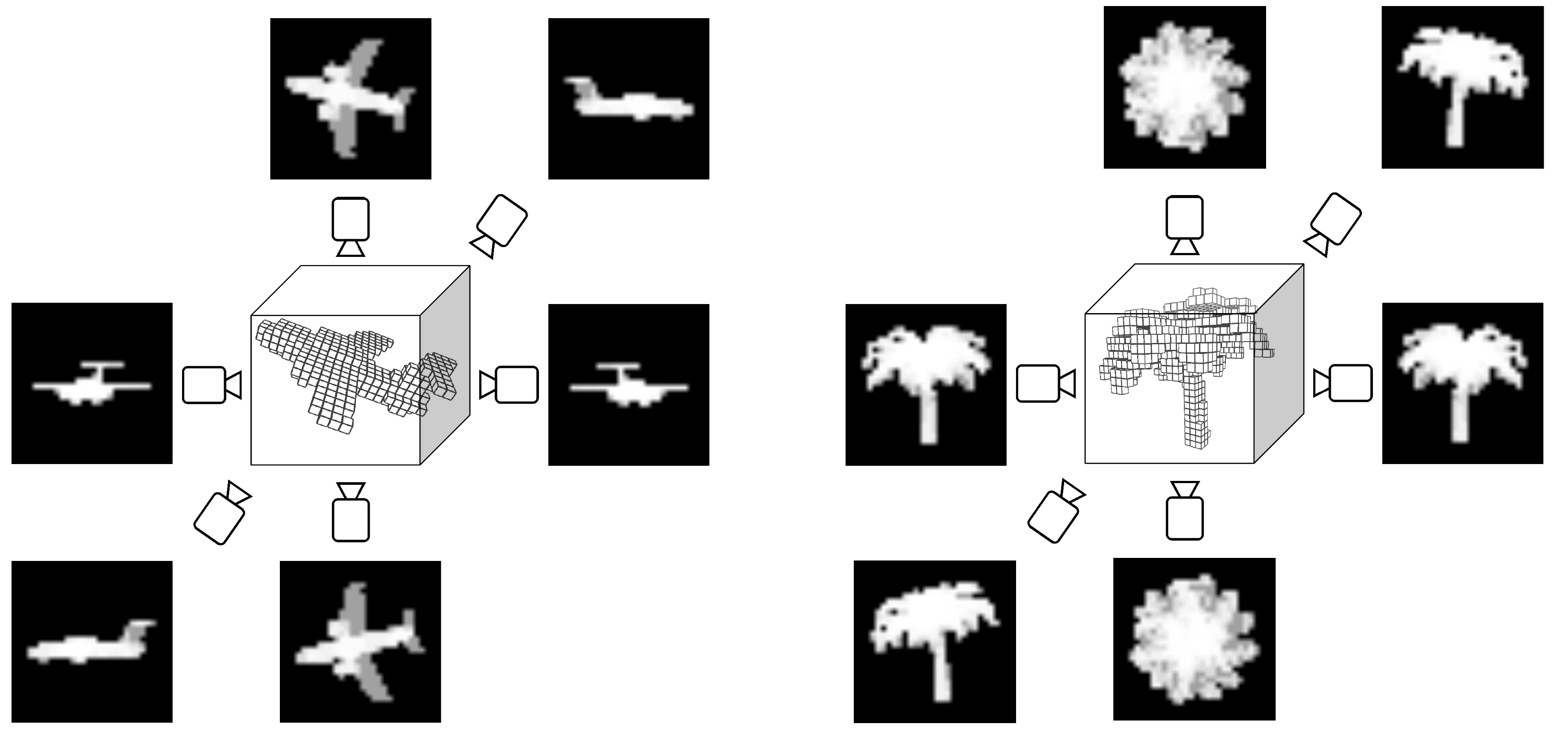}
\caption{\textbf{Examples of rendered images for the VoxMVCNN architecture}. The voxel input is rendered using the simplified technique described in Section~\ref{sec:models} to generate 6 images which are processed using the MVCNN architecture~\cite{mvcnn} for classification.}
\label{fig:voxmvcnn}
\vspace{-0.2in}
\end{figure}

We also consider a hybrid model that takes voxels as input and uses a MVCNN approach for classification using a differentiable renderer. 
To achieve this we make two simplifications. First, only six renderings are considered corresponding to viewing directions along six dimensions $(\pm x,\pm y,\pm z)$. Second, the rendering is approximated using the approach suggested in PrGAN~\cite{Gadhela:2017:PrGan} where line integrals are used to compute pixel shading color. For example the line integral of a volume occupancy grid $V$ along the axis $k$ is given by $P((i,j),V) = 1-\exp(-\sum_k V(i,j,k))$. The idea is that the higher the sum of occupancy values along the axis, the closer the integral is to 1. 
The generated views for two models are shown in Figure~\ref{fig:voxmvcnn}.
The renderings generated this way approximate silhouette renderings as described earlier.
The primary advantage of this rendering method is that it's differentiable, and hence we use this model to analyze the robustness of the MVCNN architecture to adversarial inputs (described in Section~\ref{sec:advers}).

\subsubsection{Point network (PointNet)}
We follow the same architecture as PointNet~\cite{pointnet} that operates on point cloud representations of a model.
The architecture applies a series of non-linear mappings individually to each input point and performs orderless aggregations using max-pooling operations.
Thus the model is invariant to the order in which the points are presented and can directly operate on point clouds without additional preprocessing such as spatial partitioning, or graph construction.
Additionally, some initial layers are used to perform spatial transformations (rotation, scaling, etc.)
Despite its simplicity the model and its variants have been shown to be effective at shape classification and segmentation tasks~\cite{pointnet,Qi:PointNet++:2017}.

\para{Training details.} All the MVCNN models are trained in two stages as suggested in~\cite{mvcnn}. The model is first trained as a single-image classification task where the view-pooling layer is removed, then trained to jointly classify all the views with view-pooling layer in the second stage. 
We use the Adam optimizer~\cite{kingma2014adam} with learning rate $5\times10^{-5}$ and $1\times10^{-5}$ for first and second stage respectively and each stage is trained with 30 epochs. The batch size is set to 64 and 96 (eight models with twelve views) for each stage and the weight decay parameter is set to $0.001$. 
The VoxNet is trained with Adam optimizer with learning rate $1\times10^{-3}$ for 150 epochs. The batch size is set to 64 and weight decay parameter is $0.001$. The VoxMVCNN is trained using the same procedure as MVCNN. For PointNet~\cite{pointnet} we use the publicly available implementation released by the authors.

\subsection{Generating Adversarial Inputs}
\label{sec:advers}
Adversarial examples to image-based deep neural networks have been thoroughly explored in the 
literature~\cite{intriguingnn2013,explainingadv2015,physicaladv2016,nguyen2015deep}.
However, there is no prior work that addresses adversarial examples to deep neural networks based
on 3D representations.
Here we want to investigate if adversarial shapes can be obtained from different 3D shape recognition models, 
and perhaps more importantly which 3D representation is more robust to adversarial examples. 
We define an adversarial example as follows.
Consider $\phi(\mathbf{s}, y)$ as the score that a classifier $\phi$ gives to an input $\mathbf{s}$ belonging to a class $y$.
An adversarial example $\mathbf{s^\prime}$ is a sample that is perceptually similar to $\mathbf{s}$,
but $\argmax_y \phi(\mathbf{s}, y) \neq \argmax_y \phi(\mathbf{s^\prime}, y)$.
It is known from~\cite{intriguingnn2013} that an effective way to compute adversarial examples to
image-based models is the following.
Given a sample from the dataset $\mathbf{s}$ and a class $y^\prime$ that one wishes to maximize the score,
an adversarial example $\mathbf{s^\prime}$ can be computed as follows 
\begin{equation}
	\mathbf{s^\prime} = \mathbf{s} + \alpha \nabla_{\mathbf{s}}\phi(\mathbf{s}, y^\prime)\nonumber
\end{equation}
where $\nabla_{\mathbf{s}}$ is the gradient of the classifier with respect to the input $\mathbf{s}$, and $\alpha$ is the learning rate.
For many image models, this single step procedure is able to generate perceptually indistinguishable
adversarial examples.
However, for some of the examples we experimented, a single step is not enough to generate a misclassification.
Thus, we employ the following iterative procedure based on~\cite{physicaladv2016}:
\begin{equation}
	\label{eq:advers}
	\mathbf{s^\prime_0} = \mathbf{s}, \quad
	\mathbf{s^\prime_{t+1}} = clip_{\mathbf{s}, \epsilon}
		\big\{\mathbf{s}_{t} + \alpha \nabla_{\mathbf{s}}\phi(\mathbf{s_{t}}, y^\prime)\big\}
\end{equation}
where $clip_{\mathbf{s}, \epsilon}\{x\}$ is an operator that clips the values of $x$ to make sure the
result will be in the $L_\infty$ $\epsilon$-neighborhood of $\mathbf{s}$. 
Notice that this procedure is agnostic to the representation used by the input data $\mathbf{s}$.
Thus, we use the same method to generate adversarial examples for multiple modalities of
3D representation: voxels, point clouds, multi-view images.
For voxel grids, we also clip the values of $\mathbf{s}$ to make sure their values are in
$[0, 1]$. 
For multi-view representations, we need to make sure that all views are consistent with each other.
We address this issue by using the VoxMVCNN architecture that generates multiple views from the same
object through a differentiable renderer, i.e. line integral, as described in Section~\ref{sec:VoxMVCNN}. 

\section{Experiments}
We begin by investigating the model generalization in Section~\ref{sec:few_examples}. 
Section~\ref{sec:MVCNN} analyzes the effect of different architectures and renderings for the MVCNN.
Section~\ref{sec:auxiliary} uses cross-modal distillation to improve the performance of VoxNet and PointNet.
Section~\ref{sec:representations} compares the tradeoffs between different representations.
Section~\ref{sec:adversarial} compares the robustness of different classifiers to adversarial perturbations.
Finally, Section~\ref{sec:comparison} puts the results presented in this paper in the context of prior work.

\subsection{Learning From a Few Examples}
\label{sec:few_examples}
\begin{figure}[t]
\begin{center}
\includegraphics[clip, width=0.45\linewidth]{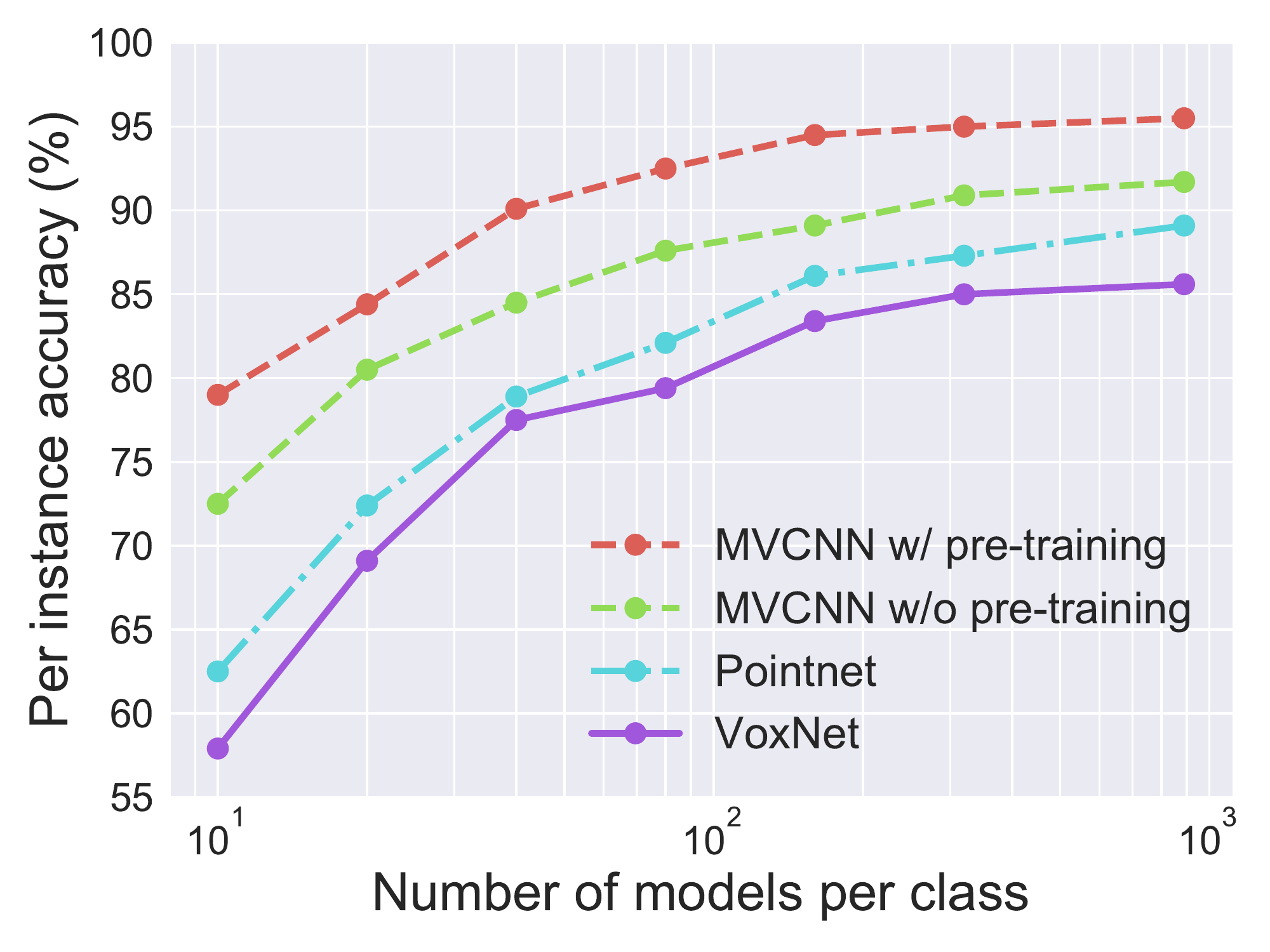}
\includegraphics[clip, width=0.45\linewidth]{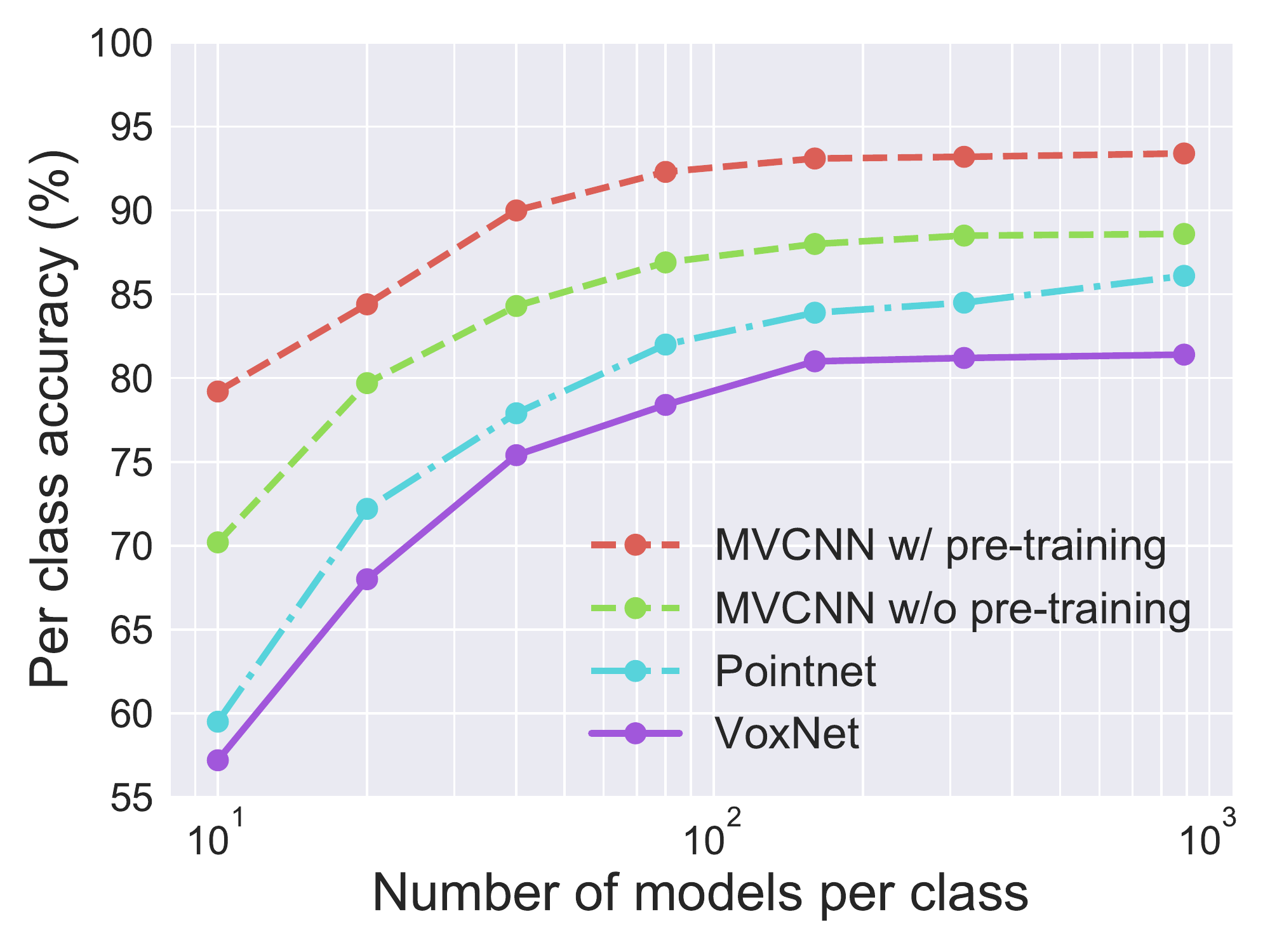}
\end{center}
\vspace{-0.3in}
\caption{Classification accuracy as a function of training set size. MVCNN generalizes better than the other approaches. The two MVCNN curves correspond to variants with and without ImageNet pretraining.}
\label{fig:few_example}
\end{figure}

One of the most desirable properties of a classifier is its ability to generalize from a few examples.
We test this ability by evaluating the accuracy of different models as a function of training set size. We select the first $M_k$ models in the training set for each class, where
\[ 
M_k=\min(N_k,\{10,20,40,80,160,320,889\}),
\]
and $N_k$ is the number of models in class $k$. 
The maximum number of models per-class in the training set of ModelNet40 is 889. 
Figure~\ref{fig:few_example} shows the per-class and per-instance accuracy for three different models as a function of the training set size.
The MVCNN with the VGG-11 architecture has better generalization than VoxNet and PointNet across all training set sizes. 
MVCNN obtains 77.8\% accuracy using only 10 training models per class, while PointNet and VoxNet obtain 62.5\% and 57.9\% respectively. 
The performance of MVCNN is near optimal with 160 models per class, far fewer than PointNet and VoxNet. 
When using the whole dataset for training, MVCNN (95.0\%) outperforms PointNet (89.1\%) and VoxNet (85.6\%) by a large margin.

Several improvements have been proposed for both point-based and voxel-based architectures. The best performing point-based models to the best of our knowledge is the Kd-Networks~\cite{Klokov_2017_ICCV} which achieves 91.8\% per-instance accuracy. 
For voxel-based models, O-CNN~\cite{Wang-2017-OCNN} uses sparse convolutions to handle higher resolution with Octave trees~\cite{meagher1980octree} and achieves 90.6\% per-instance accuracy.
However, all of them are far below the MVCNN approach. 
More details and comparison to the state-of-the-art are in Section~\ref{sec:comparison}.

\subsection{Dissecting the MVCNN Architecture}
Given the high performance of MVCNN we investigate what factors contribute to its performance as described next.

\label{sec:MVCNN}
\para{Effect of model architecture.}
The MVCNN model in~\cite{mvcnn} used VGG-M architecture. However a number of different image networks have since been proposed.
We used different CNN architectures for MVCNN and report the accuracies in Table~\ref{tab:mvcnn_architecture}. 
All models have similar performance suggesting that MVCNN is robust across different CNN architectures. 
In Table~\ref{tab:mvcnn_input} we also compare with the results using VGG-M and AlexNet. With the same shaded images and training subset, VGG-11 achieves 89.1\% and VGG-M has 89.9\% accuracy.
\begin{table}[h!]
\centering
\begin{tabular}{c|c|c}
Model & Per class & Per instance \\
\hline
VGG-11 & 92.4 & 95.0 \\
ResNet 18 & 92.8 & 95.6 \\ 
ResNet 34 & 93.4 & 95.9 \\
ResNet 50 & 94.0 & 95.5 \\
\end{tabular}
\vspace{0.1in}
\caption{Accuracy (\%) of MVCNN with different CNN architectures. The VGG-11 architectures are on par with the residual network variants. }
\vspace{-0.3in}
\label{tab:mvcnn_architecture}
\end{table}

\para{Effect of ImageNet pretraining.}
MVCNN benefits from transfer learning from ImageNet classification task.
However, even without ImageNet pretraining, the MVCNN achieves 91.3\% per-instance accuracy (Table~\ref{tab:pretraining}). 
This is higher than several point-based and voxel-based approaches.
Figure~\ref{fig:few_example} plots the performance of the MVCNN with VGG-11 network without ImageNet pretraining across training set sizes showing this trend is true throughout the training regime.
In Section~\ref{sec:auxiliary} we study if ImageNet pretraining can benefit such approaches using cross-modal transfer learning.


\begin{table}[h!]
\centering
\vspace{-0.1in}
\begin{tabular}{c|c|c}
Model & Per class & Per instance \\
\hline
VGG-11 w/~ ImageNet pretraining & 92.4 & 95.0 \\
VGG-11 w/o ImageNet pretraining & 88.7 & 91.3
\end{tabular}
\vspace{0.1in}
\caption{Effect of ImageNet pretraining on the accuracy (\%) of MVCNN. The VGG-11 architecture is used and the full training/test split of the ModelNet40 dataset is used.}
\label{tab:pretraining}
\vspace{-0.4in}
\end{table}

\para{Effect of shape rendering.}
We analyze the effect of different rendering approaches for input to a MVCNN model in Table~\ref{tab:mvcnn_input}. Sphere rendering proposed in~\cite{volumetriccnn} refers to rendering each point as a sphere and was shown to improve performance with AlexNet MVCNN architectures.
We first compared the tight field-of-view rendering with black background in this work to the rendering in~\cite{mvcnn}.
Since~\cite{mvcnn} only reported results on the 80/20 training/test split, we first compared the performance of the VGG-11 networks using images from~\cite{mvcnn}.
The performance difference was negligible.
However with our shaded images the performance of the VGG-11 network improves by more than 2\%.

Using depth images, the per instance accuracy is 3.4\% lower than using shaded images, but concatenating shaded images with depth images gives 1.2\% improvement.
Furthermore, we found the shading information only provides 1.4\% improvements over the binary silhouette images.
This suggests that most of the discriminative shape information used by the MVCNN approaches lie in the boundary of the object.

\begin{table}[h!]
\centering
\begin{tabular}{c|c|c|c||c|c}
& & \multicolumn{2}{c||}{Full training/test} & \multicolumn{2}{|c}{80/20 training/test}\\
Model & Rendering & Per class & Per instance & Per class & Per instance\\
\hline
VGG-M & Shaded from~\cite{mvcnn} & - & - & 89.9 & 89.9\\
VGG-M & Shaded from~\cite{mvcnn} (80$\times$) & - & - & 90.1 & 90.1\\
\hline
VGG-11 & Shaded from~\cite{mvcnn}& - & - & 89.1 & 89.1 \\
VGG-11 & Shaded & {\bf 92.4} & {\bf 95.0} &  {\bf 92.4} & {\bf 92.4} \\
VGG-11 & Depth & 89.8 & 91.6 \\ 
VGG-11 & Shaded + Depth & 94.7 & 96.2 \\
VGG-11 & Silhouettes & 90.7 & 93.6 \\
\hline
AlexNet & Sphere rendering (20$\times$) & 89.7 & 	92.0 \\ 
AlexNet-MR & Sphere rendering (20$\times$) & 91.4 & 93.8 \\
\end{tabular}
\vspace{0.2in}
\caption{Accuracy (\%) of MVCNN with different rendering methods. The number in the brackets are the number of views used. 12 views are used if not specified.}
\label{tab:mvcnn_input}
\vspace{-0.5in}
\end{table}

%


\subsection{Cross Modal Distillation}
\label{sec:auxiliary}
%
%

Knowledge distillation~\cite{Caruana06,Hinton14} was proposed for model compression tasks. 
They showed the performance of the model can be improved by training to imitate the output of a more accurate model. This technique has also been applied on transferring rich representations across modalities. 
For example, a model trained with images can be used to guide learning of a model for depth images~\cite{Gupta_CVPR2016}, or to a model for sound waves~\cite{aytar2016soundnet}. 
We investigate such techniques for learning across different 3D representations; Specifically from MVCNN model to PointNet and VoxNet models.

\begin{figure}[h!]
\begin{center}
\includegraphics[clip, width=0.45\linewidth]{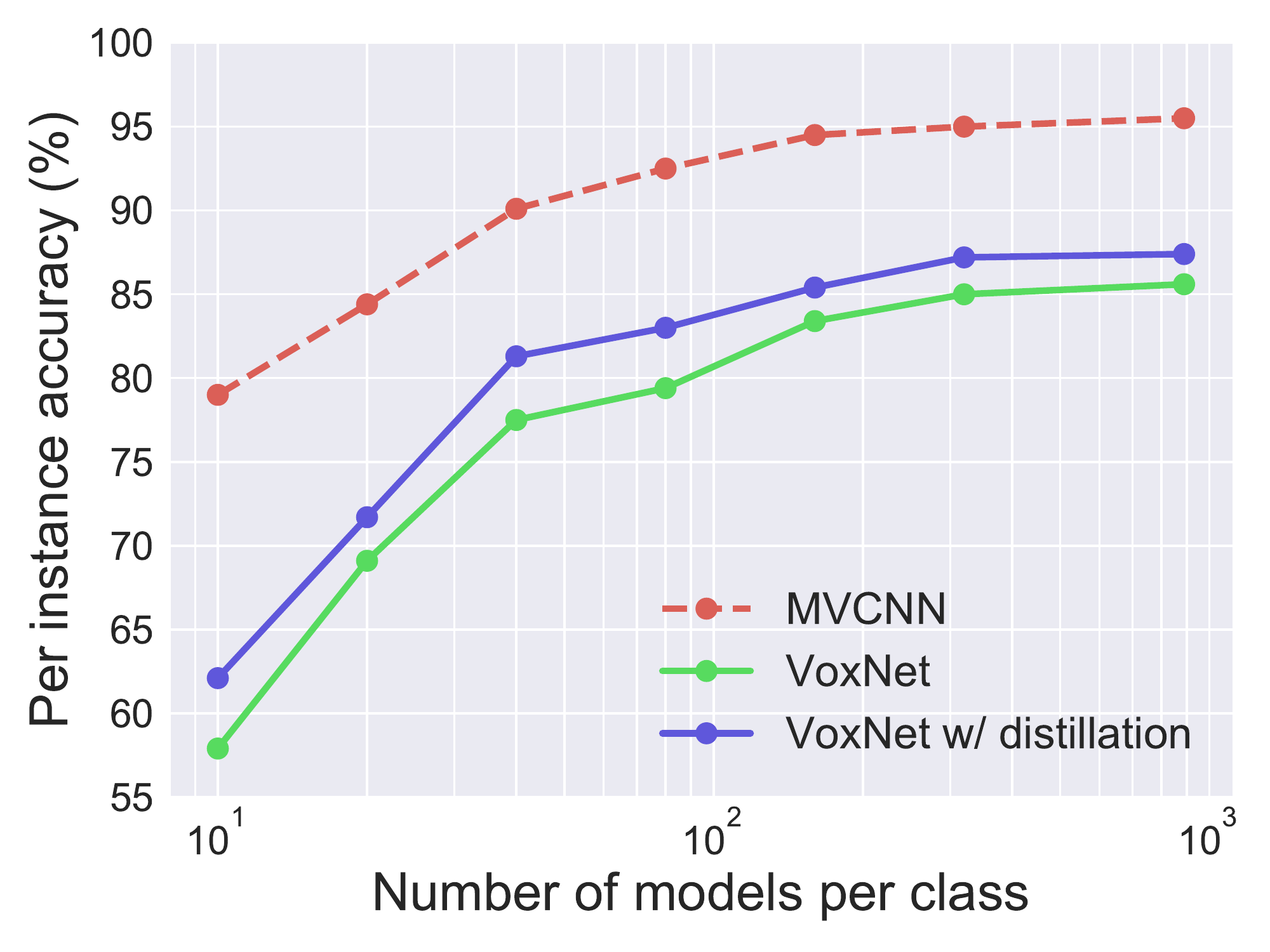}
\includegraphics[clip, width=0.45\linewidth]{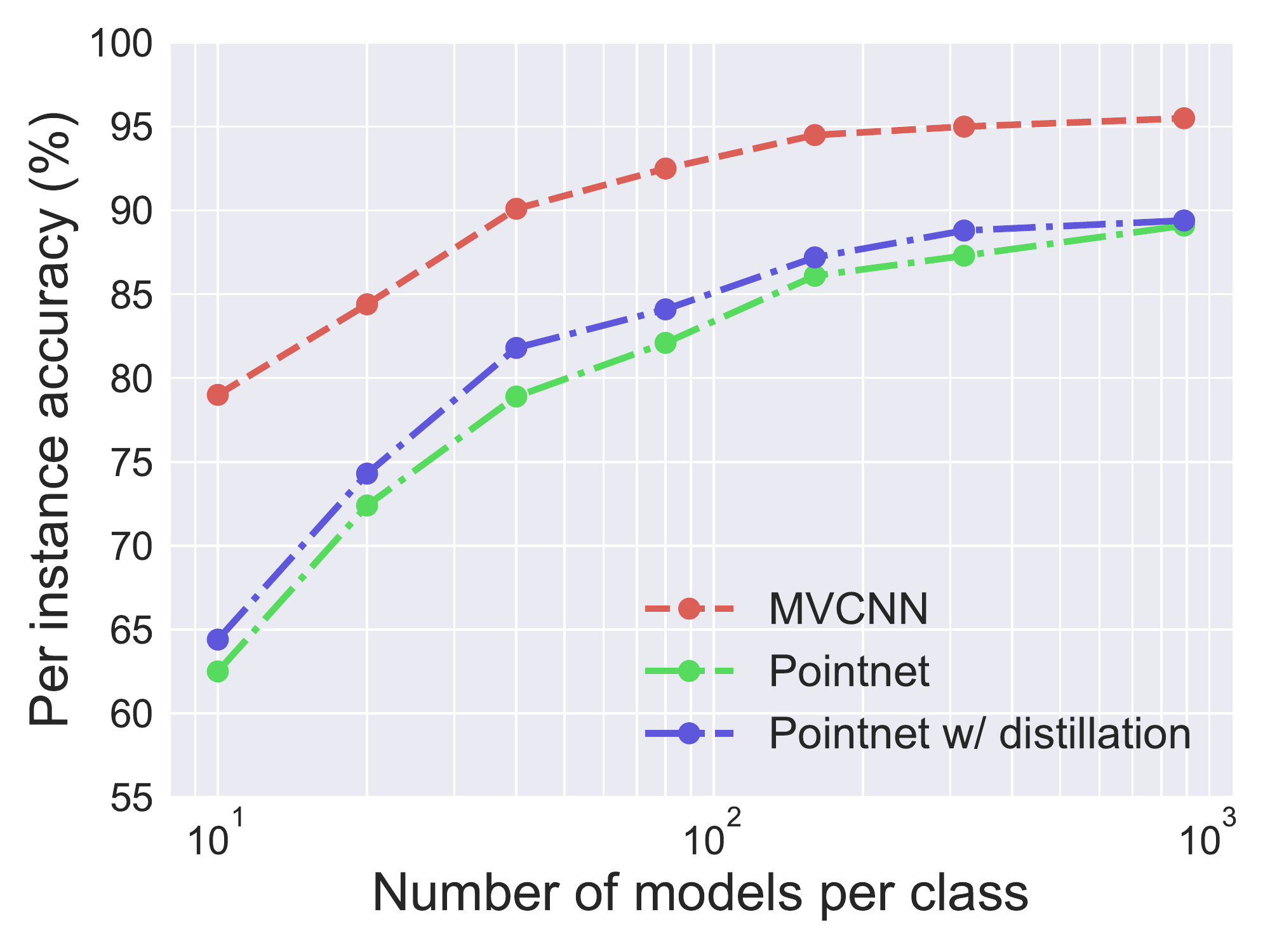}
\end{center}
\vspace{-0.2in}
\caption{Model distillation from MVCNN to VoxNet and PointNet. The accuracy is improved by 1.8\% for VoxNet and 0.3\% for PointNet with whole training set. }
\label{fig:distillation}
\end{figure}

To do this we first train the ImageNet initialized VGG-11 network on the full training set. The logits (the last layer before the softmax) are extracted on the training set. 
A PointNet (or VoxNet) model is then trained to minimize
\begin{equation}\label{eqn:cqd}
 \sum_{i=1}^{n} 
 	{\cal L}\left( \sigma({\bf z}_i), y_i \right) + 
 	\lambda \sum_{i=1}^{n} {\cal L}
 	\left(
 	\sigma\left(
 			\frac{{\bf z}_i}{T}
 			\right), 
 	\sigma\left( 
	 		\frac{{\bf x}_i}{T}
 			\right)
 	\right) 
\end{equation}
where ${\bf x}_i$ and ${\bf z}_i$ are the logits from the MVCNN model and from the model being trained respectively, $y_i$ is the class label of the input $\mathbf{s}_i$, $\sigma(x)$ is the softmax function, and ${\cal L}$ is the cross-entropy loss ${\cal L}(p,q)=-\sum{p_i\log q_i}$.
$T$ is the temperature for smoothing the targets. 
$\lambda,T$ are set by grid search for $T\in[1,20],\lambda\in[1,100]$. 
For example, in PointNet the best hyper-parameters are $T=20,\lambda=50$ when training set is small, and $T=15,\lambda=10$ when the training set is larger. 
In VoxNet we set $T=10,\lambda=100$ in all cases.
Figure~\ref{fig:distillation} shows the result of training VoxNet and PointNet with distillation. 
For VoxNet the per instance accuracy is improved from 85.6\% to 87.4\% with whole training set; 
For PointNet the accuracy is improved from 89.1\% to 89.4\%. 
The improvement is slightly bigger when there is less training data. 

\vspace{-0.3in}

\subsection{Tradeoffs between Learned Representations}
\label{sec:representations}
In this Section we analyze the tradeoffs between the different shape classifiers. 
Table~\ref{tab:speed} compares their speed, memory, and accuracy. 
The MVCNN model has more parameters and is slower, but the accuracy is 5.9\% better than PointNet and 9.4\% better than VoxNet.
Even though the number of FLOPS are far higher for MVCNN the relative efficiency of 2D convolutions results in slightly longer evaluation time compared to VoxNet and PointNet.

\begin{figure}[h!]
\begin{center}
\includegraphics[clip, width=0.45\linewidth]{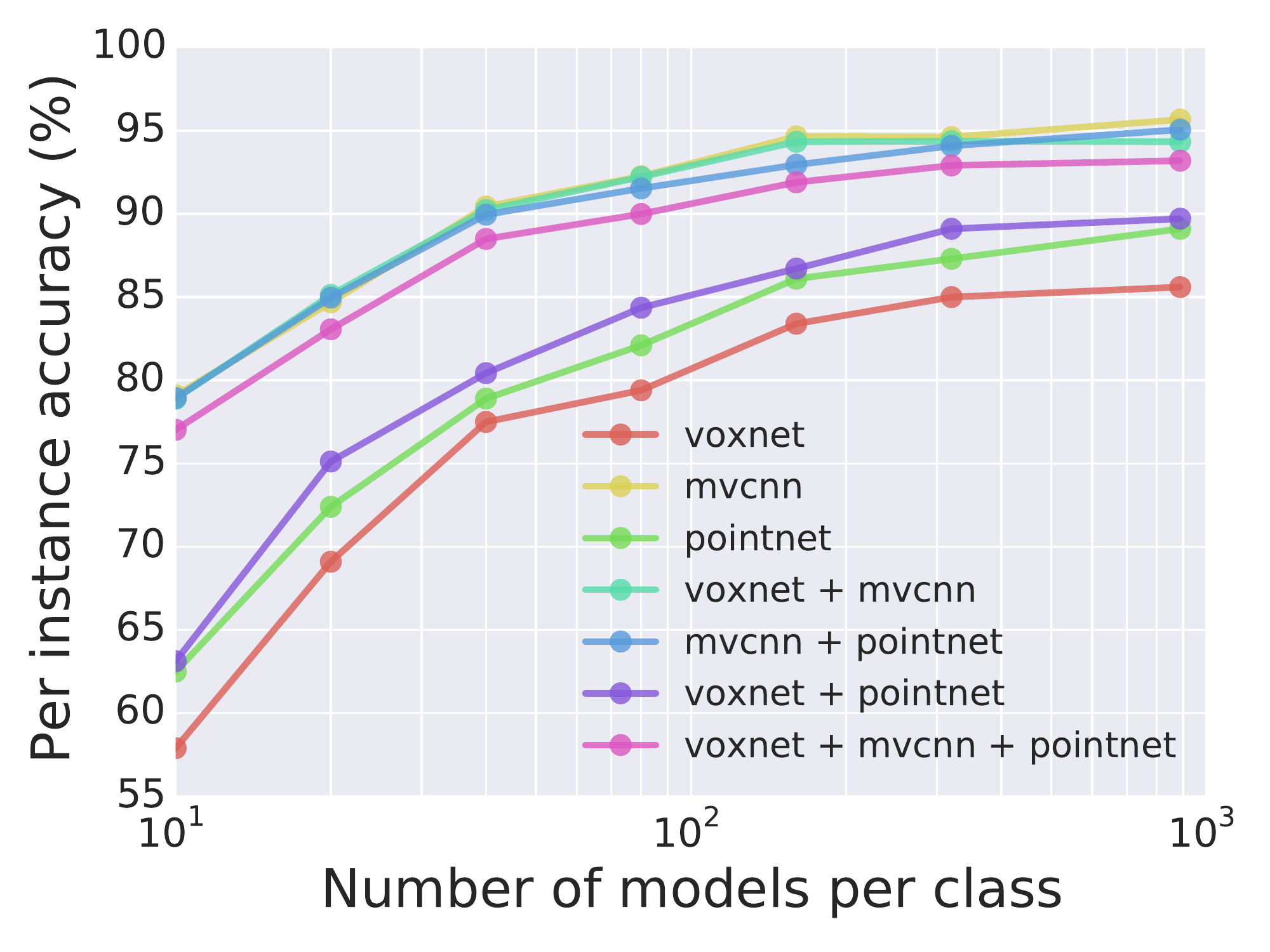}
\includegraphics[clip, width=0.45\linewidth]{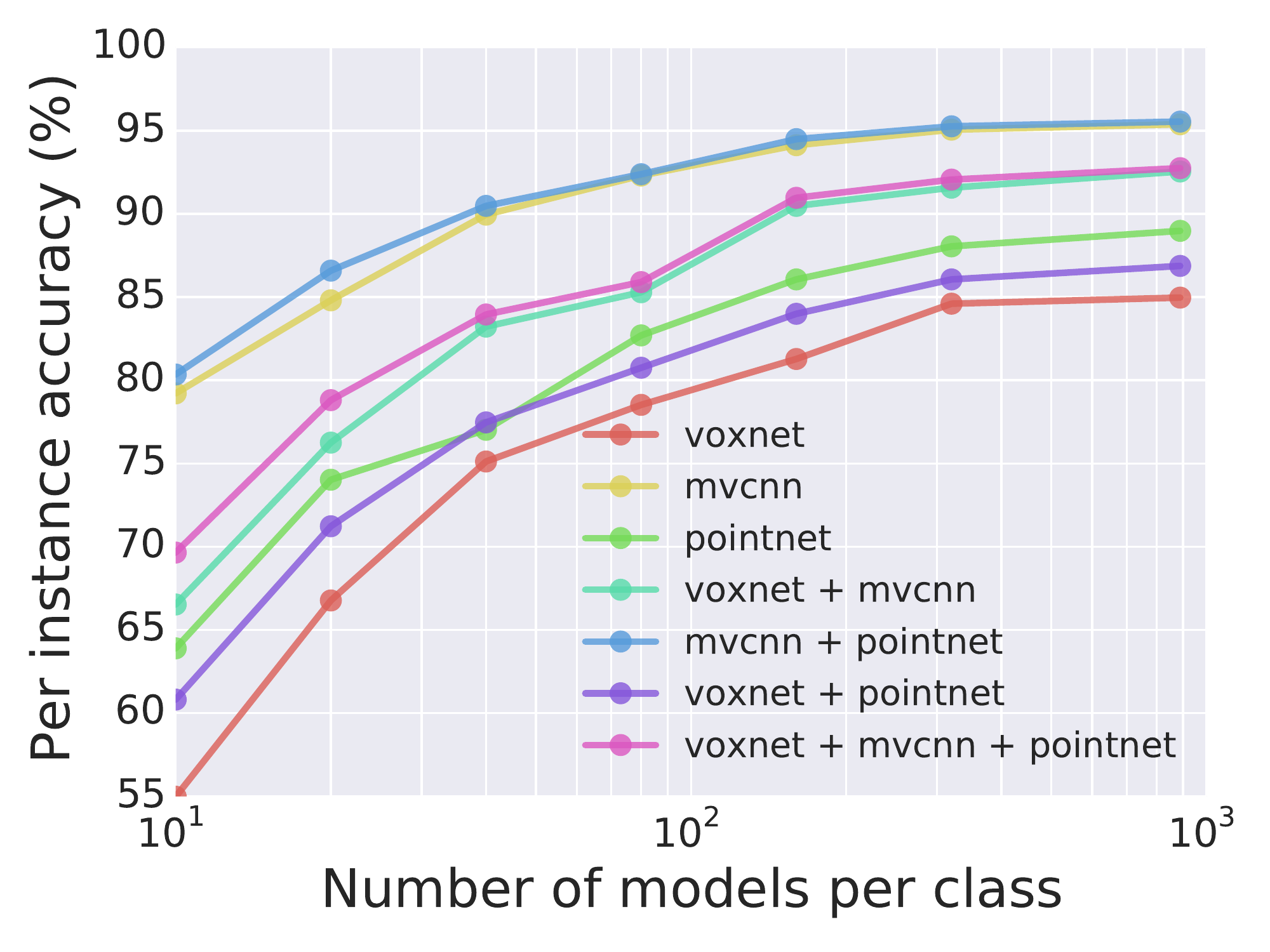}
\end{center}
\vspace{-0.2in}
\caption{Accuracy obtained by ensembling models. Left shows results by averaging the predictions while right shows results by training a linear model on the concatenated features extracted from different models.}
\vspace{-0.2in}
\label{fig:concat_feat}
\end{figure}

We further use an ensemble model combining images, voxels, and point cloud representations. 
A simple way is to average the predictions from different models. As shown in Figure~\ref{fig:concat_feat}, the ensemble of VoxNet and PointNet has better performance than using single model. However, the predictions from MVCNN dominate VoxNet and PointNet and gives no benefit for combining the predictions from other models with MVCNN.
A more complex scheme where we trained a linear model on top of features extracted from penultimate layers of these networks did not provide any improvements either.

\begin{table}[h!]
\centering
\begin{tabular}{c||c|c|c|c|c}
Model & Forward-pass time & \#params & Memory (GB) & Per class acc. (\%) & Per ins. acc. (\%)\\
\hline
MVCNN & 25.8 ms & 128.9M & 10.0 & 92.4 & 95.0 \\
VoxNet & 1.3 ms & 1.4M & 2.0 & 81.4 (82.5) & 85.6 (87.4) \\
PointNet & 3.1 ms & 3.5M & 4.4 & 86.1 (86.7) & 89.1 (89.4) 
\end{tabular}
\vspace{0.1in}
\caption{Accuracy, speed and memory comparison of different models. 
Memory usage during training which includes parameters, gradients, and layer activations, for a batch size 64 is shown. Forward-pass time is also calculated with batch size 64 using PyTorch with a single GTX Titan X for all the models. The input resolutions are $224\times224$ for MVCNN, $32^3$ for VoxNet, and $1024$ points for PointNet.
The accuracy numbers in brackets are for models trained with distillation as described in Section~\ref{sec:auxiliary}.}
\label{tab:speed}
\end{table}

%

\subsection{Robustness to Adversarial Examples}
\label{sec:adversarial}
In this section we analyze and compare the robustness of three shape classification models to adversarial examples. 
Adversarial examples are generated using the stochastic gradient ascent procedure described in Section~\ref{sec:advers}. 
We search the threshold $\epsilon$ from $0.001$ to $0.9$, and find the minimum value of $\epsilon$ where we can generate an 
adversarial example in 1000 iterations with learning rate $\alpha = 1\times 10^{-6}$. 

To make a quantitative analysis between VoxMVCNN and VoxNet, we use the following procedure. 
Given an input $\mathbf{s}$ and a classifier $\phi(\mathbf{s}, y)$, 
the ``hardest'' target class is defined as $y^\prime_{\mathbf{s},\phi(\mathbf{s}, y)} = \argmin_y \phi(\mathbf{s}, y)$. 
We select the first five models for each class from the test set. 
For each model, we select two target classes $y^\prime_{\mathbf{s},\phi_1(\mathbf{s}, y)}$ and $y^\prime_{\mathbf{s},\phi_2(\mathbf{s}, y)}$ where $\phi_1$ is VoxMVCNN and $\phi_2$ is VoxNet. There are total 400 pairs of test cases $(\mathbf{s},y^\prime)$. 
We say an adversarial example $\mathbf{s^\prime}$ can be found when $y^\prime = \argmax_y \phi(\mathbf{s^\prime}, y)$ with $\epsilon\leq 0.9$.

As shown in Table~\ref{tab:epsilon}, we can generate 399 adversarial examples out of 400 test cases for VoxMVCNN, but only 370 adversarial examples for VoxNet. 
We then report the minimum $\epsilon$ where adversarial examples can be found in each test case. The average $\epsilon$ of VoxMVCNN is smaller than VoxNet, which suggests that VoxMVCNN is easier to be fooled by adversarial examples than VoxNet. We also use the VoxMVCNN model trained without ImageNet pre-training. Surprisingly, the model without pre-training is more robust to adversarial examples as the mean of $\epsilon$ is bigger than the model with pre-training.  
As for PointNet, we can generate 379 adversarial examples. 
Note that the $\epsilon$ value here is not comparable with other voxel representations, 
since in this case we are changing point coordinates instead of occupancy levels.

\begin{figure}[ht!]
  \begin{tabular}{ccccc}
Point Cloud & PointNet & Voxels & VoxNet & VoxMVCNN \\
      \parbox[c]{7.2em}{\reflectbox{\includegraphics[trim={3.5cm 3.5cm 3.5cm 3.5cm},clip,width=0.95\linewidth]{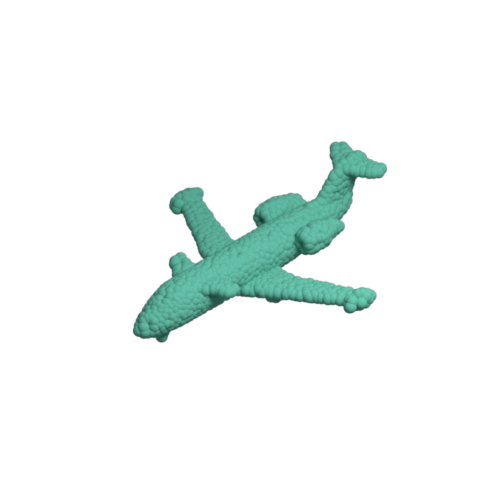}}} & 
      \parbox[c]{7.2em}{\reflectbox{\includegraphics[trim={3.5cm 3.5cm 3.5cm 3.5cm},clip,width=0.95\linewidth]{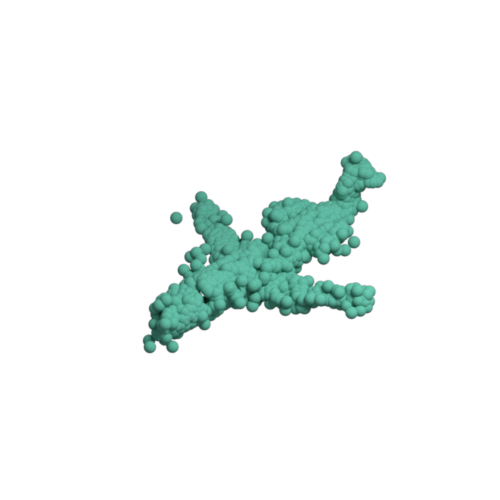}}} & 
      \parbox[c]{7.2em}{\includegraphics[trim={2.5cm 4cm 2.5cm 4cm},clip,width=0.95\linewidth]{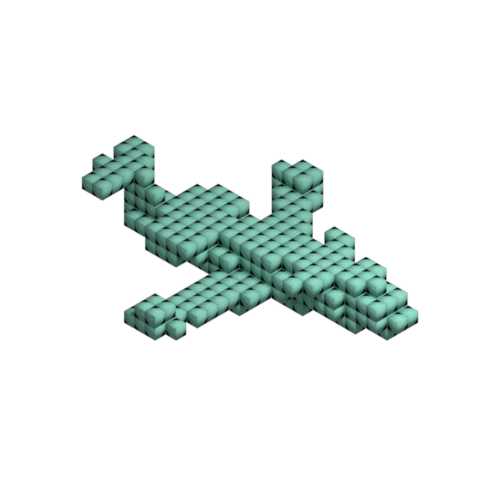}} & 
      \parbox[c]{7.2em}{\includegraphics[trim={3.3cm 4cm 3.3cm 4cm},clip,width=0.95\linewidth]{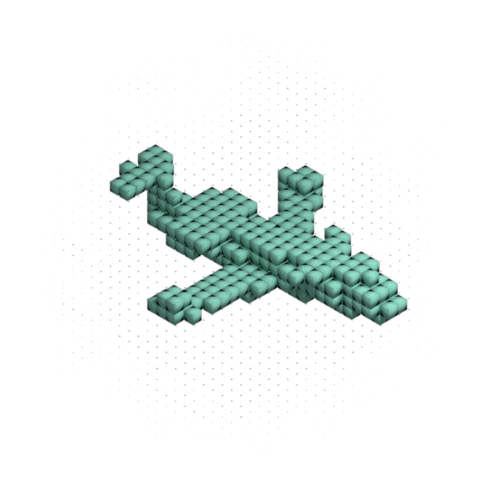}} & 
      \parbox[c]{7.2em}{\includegraphics[trim={3.3cm 4cm 3.3cm 4cm},clip,width=0.95\linewidth]{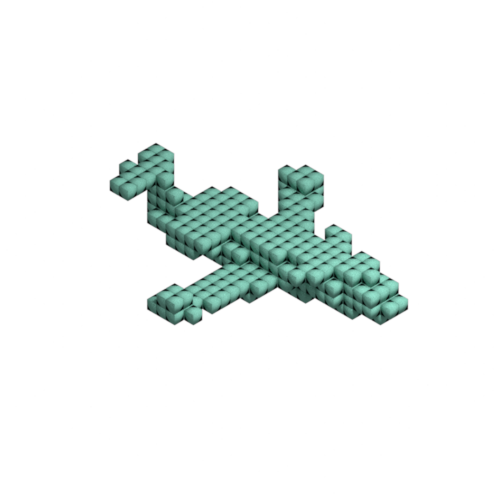}} \\
 & $\epsilon=0.05$ &   & $\epsilon=0.07$ & $\epsilon=0.003$  \\
	\parbox[c]{7.2em}{\reflectbox{\includegraphics[trim={3.5cm 3.5cm 3.5cm 3.5cm},clip,width=0.95\linewidth]{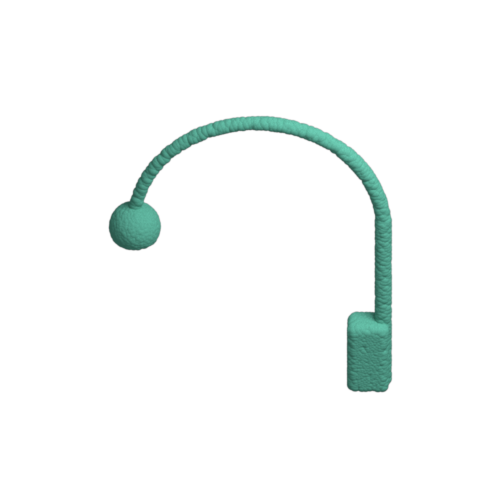}}} & 
      \parbox[c]{7.2em}{\reflectbox{\includegraphics[trim={4.5cm 4.5cm 4.5cm 4.5cm},clip,width=0.95\linewidth]{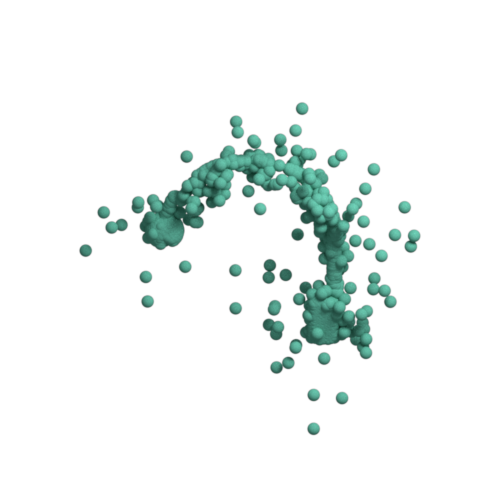}}} & 
      \parbox[c]{7.2em}{\includegraphics[trim={2.5cm 2cm 2.5cm 2cm},clip,width=0.95\linewidth]{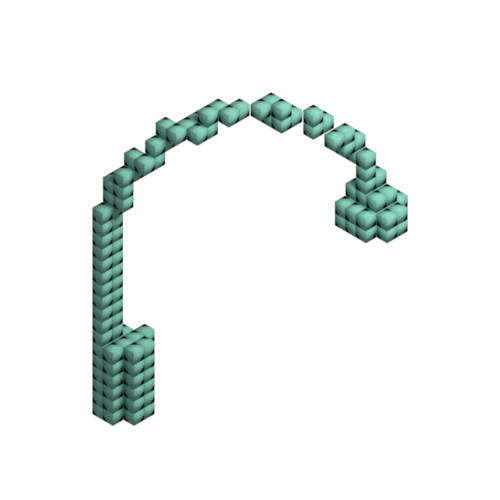}} & 
      \parbox[c]{7.2em}{\includegraphics[trim={3.3cm 3cm 3.3cm 3cm},clip,width=0.95\linewidth]{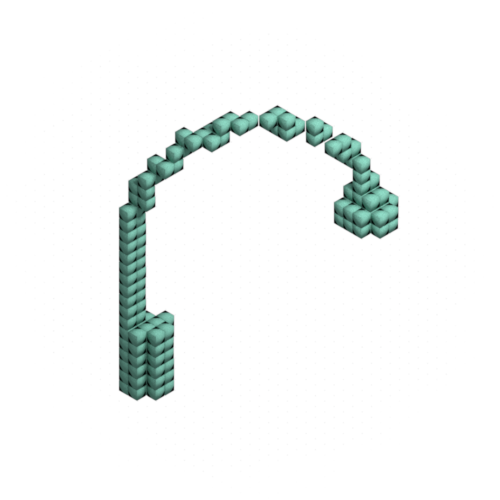}} & 
      \parbox[c]{7.2em}{\includegraphics[trim={3.3cm 3cm 3.3cm 3cm},clip,width=0.95\linewidth]{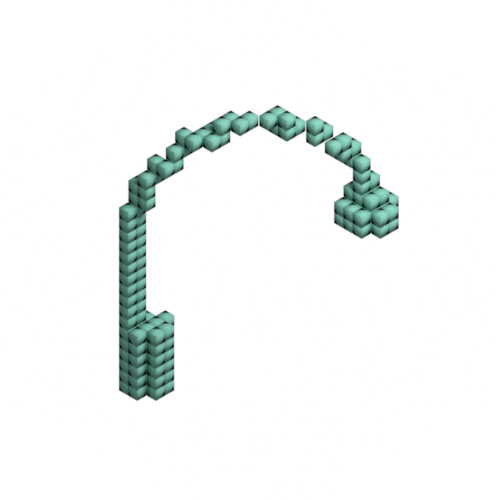}} \\
 & $\epsilon=0.4$ &   & $\epsilon=0.02$ & $\epsilon=0.004$  \\
	\parbox[c]{7.2em}{\reflectbox{\includegraphics[trim={3.5cm 3.0cm 3.5cm 3.0cm},clip,width=0.95\linewidth]{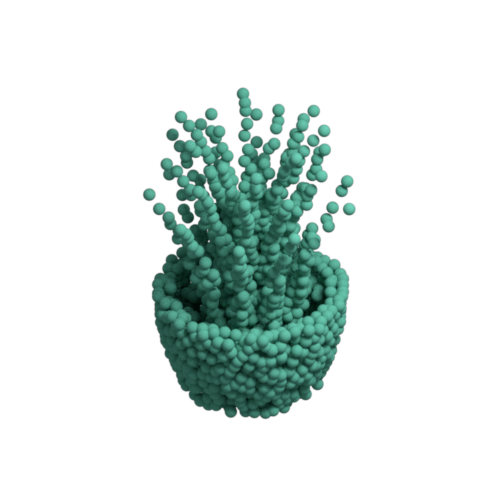}}} & 
      \parbox[c]{7.2em}{\reflectbox{\includegraphics[trim={3.8cm 2.5cm 3.8cm 2.5cm},clip,width=0.95\linewidth]{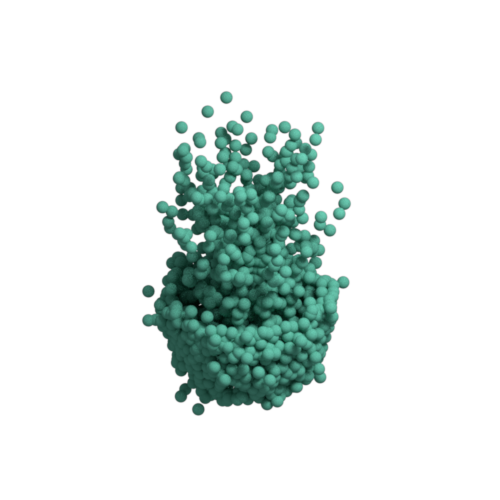}}} & 
      \parbox[c]{7.2em}{\includegraphics[trim={2.5cm 1.7cm 2.5cm 1.7cm},clip,width=0.95\linewidth]{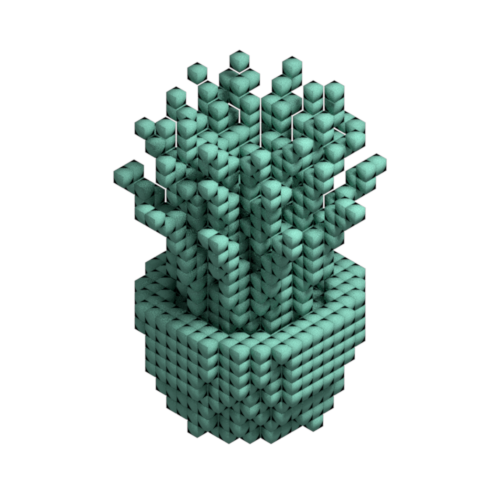}} & 
      \parbox[c]{7.2em}{\includegraphics[trim={3.3cm 2.5cm 3.3cm 2.5cm},clip,width=0.95\linewidth]{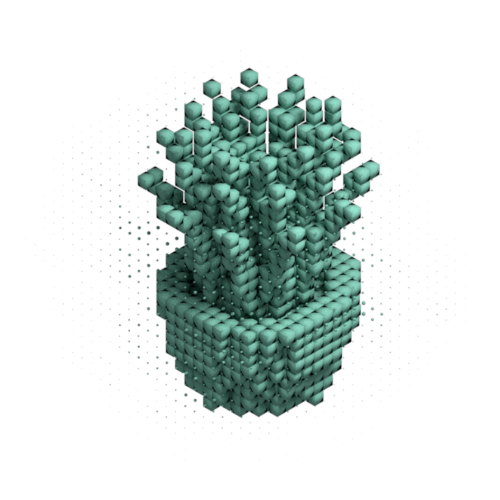}} & 
      \parbox[c]{7.2em}{\includegraphics[trim={3.3cm 2.5cm 3.3cm 2.5cm},clip,width=0.95\linewidth]{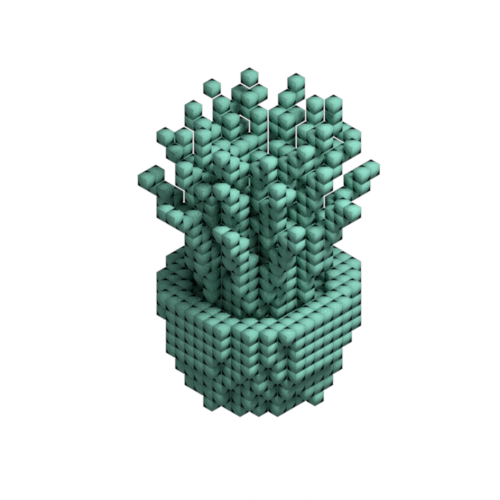}} \\
 & $\epsilon=0.04$ &   & $\epsilon=0.2$ & $\epsilon=0.007$  \\
	\parbox[c]{7.2em}{\reflectbox{\includegraphics[trim={3.5cm 3.0cm 3.5cm 3.0cm},clip,width=0.95\linewidth]{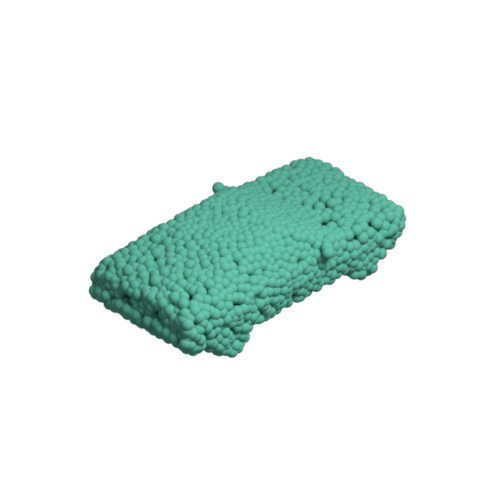}}} & 
      \parbox[c]{7.2em}{\reflectbox{\includegraphics[trim={3.8cm 2.5cm 3.8cm 2.5cm},clip,width=0.95\linewidth]{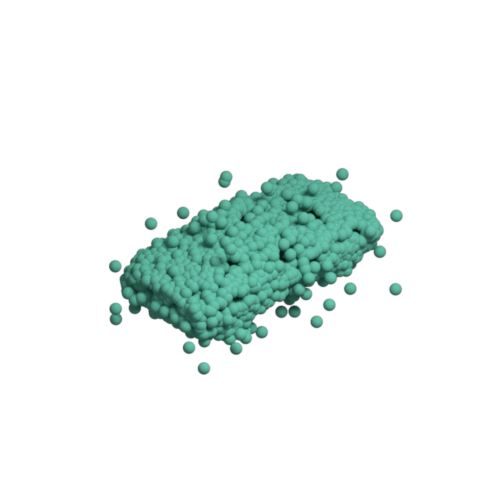}}} & 
      \parbox[c]{7.2em}{\includegraphics[trim={2.5cm 1.7cm 2.5cm 1.7cm},clip,width=0.95\linewidth]{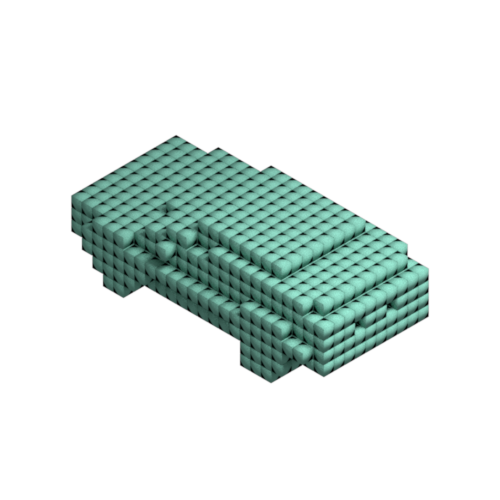}} & 
      \parbox[c]{7.2em}{\includegraphics[trim={3.1cm 2.5cm 3.1cm 2.5cm},clip,width=0.95\linewidth]{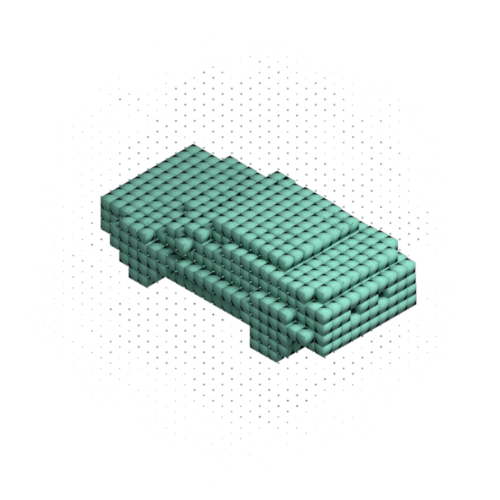}} & 
      \parbox[c]{7.2em}{\includegraphics[trim={3.1cm 2.5cm 3.1cm 2.5cm},clip,width=0.95\linewidth]{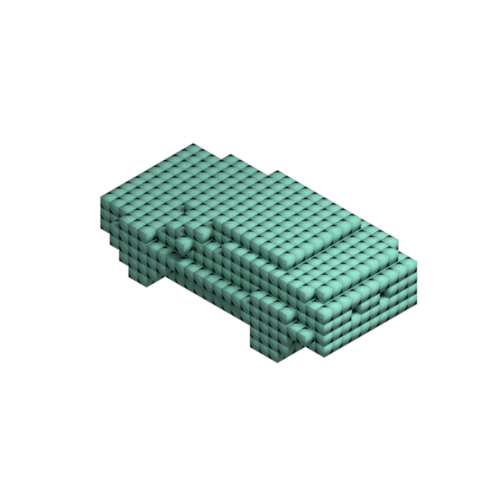}} \\
 & $\epsilon=0.2$ &   & $\epsilon=0.1$ & $\epsilon=0.02$  \\
  \end{tabular}
  \caption{Adversarial examples of PointNet, VoxNet, and VoxMVCNN. The shapes are misclassified as cup, keyboard, bench, and door for each row from top to bottom. Voxel size represents occupancy level.}
\label{fig:gallery}
\vspace{-0.2in}
\end{figure}

\begin{figure}[ht!]
  \begin{tabular}{ccccc}
&$y^\prime=$Plant & $y^\prime=$Monitor & $y^\prime=$Desk & $y^\prime=$Door\\
PointNet&
      \parbox[c]{7.2em}{\reflectbox{\includegraphics[trim={1.5cm 1cm 1.5cm 1cm},clip,width=0.95\linewidth]{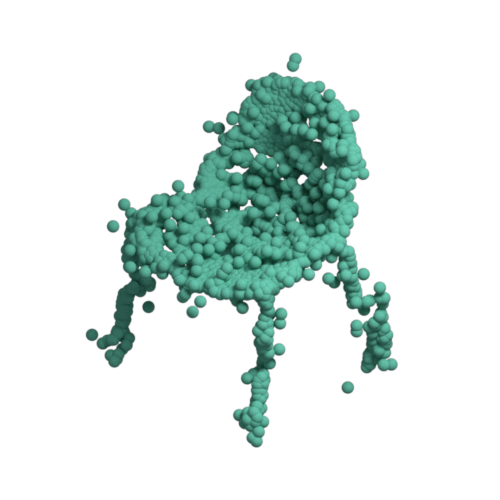}}} & 
      \parbox[c]{7.2em}{\reflectbox{\includegraphics[trim={1.5cm 1cm 1.5cm 1cm},clip,width=0.95\linewidth]{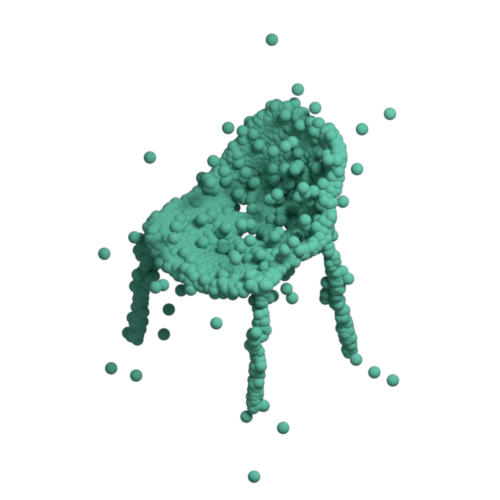}}} & 
      \parbox[c]{7.2em}{\reflectbox{\includegraphics[trim={1.5cm 1cm 1.5cm 1cm},clip,width=0.95\linewidth]{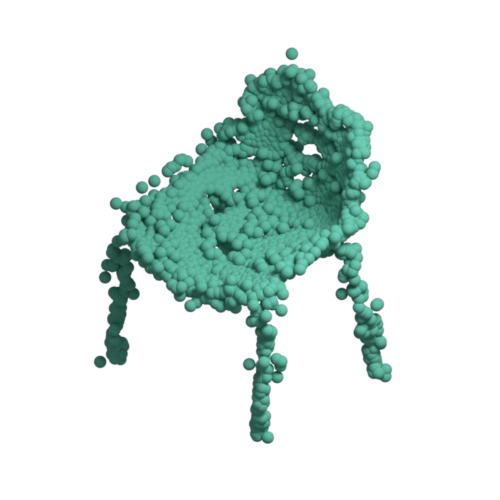}}} & 
	  \parbox[c]{7.2em}{\reflectbox{\includegraphics[trim={1.5cm 1cm 1.5cm 1cm},clip,width=0.95\linewidth]{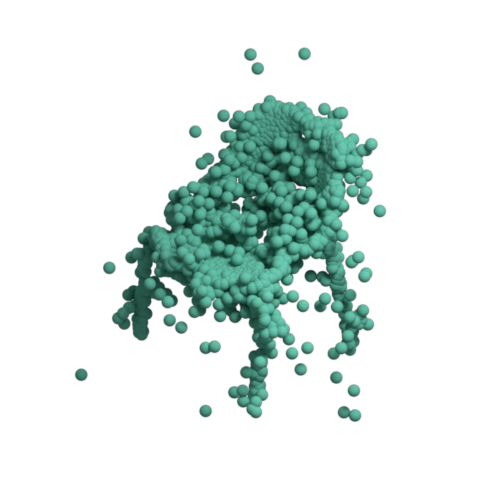}}} \\
&$\epsilon=0.08$ & $\epsilon=0.2$ &  $\epsilon=0.04$ & $\epsilon=0.2$  \\
VoxNet&
	 \parbox[c]{7.2em}{\includegraphics[trim={1.5cm 1cm 1.5cm 1cm},clip,width=0.95\linewidth]{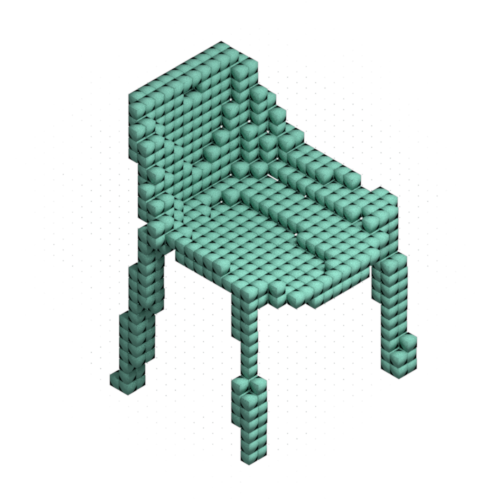}} & 
      \parbox[c]{7.2em}{\includegraphics[trim={1.5cm 1cm 1.5cm 1cm},clip,width=0.95\linewidth]{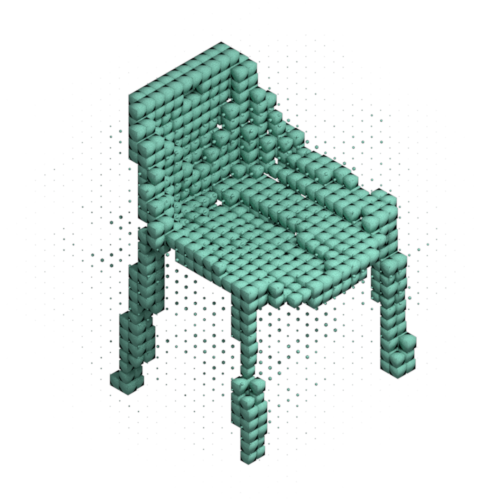}} & 
      \parbox[c]{7.2em}{\includegraphics[trim={1.5cm 1cm 1.5cm 1cm},clip,width=0.95\linewidth]{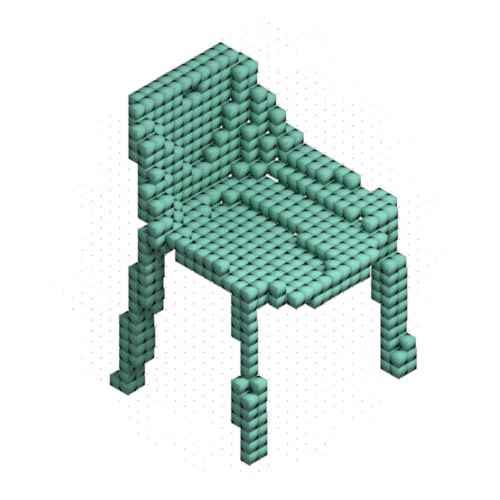}} & 
      \parbox[c]{7.2em}{\includegraphics[trim={1.5cm 1cm 1.5cm 1cm},clip,width=0.95\linewidth]{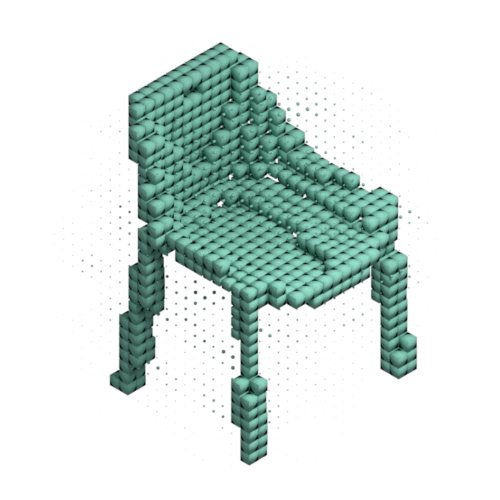}} \\
&$\epsilon=0.04$ & $\epsilon=0.2$ &  $\epsilon=0.06$ & $\epsilon=0.2$  \\
VoxMVCNN&
	  \parbox[c]{7.2em}{\includegraphics[trim={1.5cm 1cm 1.5cm 1cm},clip,width=0.95\linewidth]{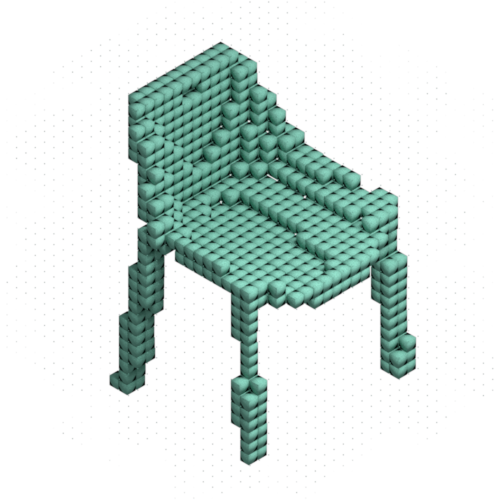}} &
	  \parbox[c]{7.2em}{\includegraphics[trim={1.5cm 1cm 1.5cm 1cm},clip,width=0.95\linewidth]{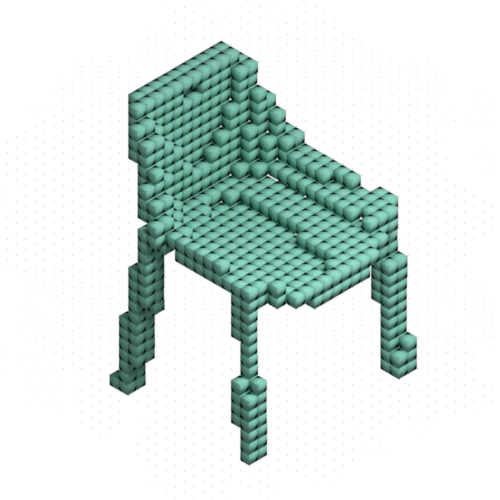}} &
	  \parbox[c]{7.2em}{\includegraphics[trim={1.5cm 1cm 1.5cm 1cm},clip,width=0.95\linewidth]{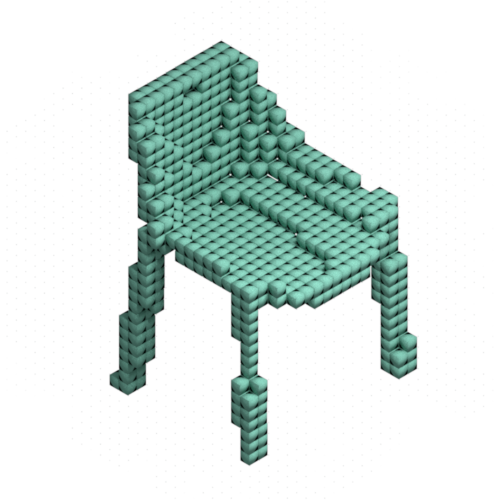}} &
      \parbox[c]{7.2em}{\includegraphics[trim={1.5cm 1cm 1.5cm 1cm},clip,width=0.95\linewidth]{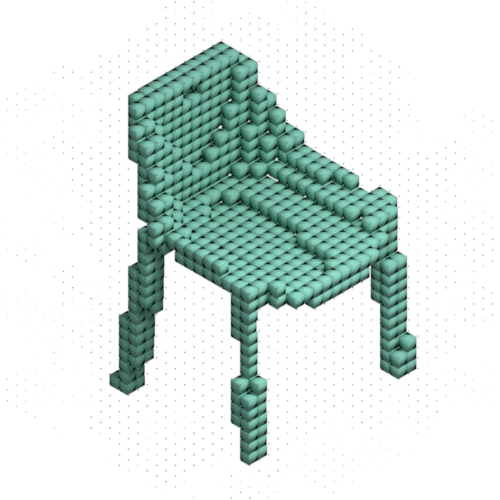}} \\
&$\epsilon=0.05$ & $\epsilon=0.04$ &  $\epsilon=0.03$ & $\epsilon=0.07$  \\
  \end{tabular}
  \caption{Adversarial examples of PointNet, VoxNet, and VoxMVCNN. The shapes are misclassified as plant, monitor, desk, and door for each column from left to right.}
\vspace{-0.1in}
\label{fig:gallery2}
\end{figure}

We also show some qualitative adversarial examples in Figure~\ref{fig:gallery} and Figure~\ref{fig:gallery2}. 
The voxels are scaled according to their occupancy level.
In Figure~\ref{fig:gallery} the input and the target classes are the same for each row, where the target labels are cup, keyboard, bench, and door from top to bottom. 
In Figure~\ref{fig:gallery2} we use the same input model but set different target class for each column.
The differences of adversarial examples for VoxMVCNN are almost imperceptible, as the $\epsilon$ is too small and the model is easy to be fooled. 
The classification accuracy of each model is shown in Table~\ref{tab:epsilon} as reference. 

\begin{table}[h!]
\vspace{-0.2in}
\centering
\begin{tabular}{c||c|c|c|c}
 & PointNet & VoxNet & \begin{tabular}{@{}c@{}}VoxMVCNN \\ w/o pre-training\end{tabular} & \begin{tabular}{@{}c@{}}VoxMVCNN \\ w/ pre-training\end{tabular} \\
\hline
\# Adversarial examples & 379 & 370 & 290 & 399 \\
\hline
$\epsilon$ & 0.045 $\pm$ 
0.054 & 0.061 $\pm$ 0.057 & 0.041 $\pm$ 0.027 & 0.006 $\pm$ 0.006\\
\hline
Per inst. acc. (\%) & 86.8 & 85.6 & 84.4 & 88.2 \\
\end{tabular}
\vspace{0.05in}
\caption{Robustness of three models to adversarial examples. 
	We generate adversarial examples with 400 test cases $(\mathbf{s},y^\prime)$ for each model. 
	The $\epsilon$ defines a $L_\infty$ $\epsilon$-neighborhood of the values that are either point coordinates or voxel occupancy. 
	We report the average of the minimum $\epsilon$ where an adversarial example can be found. 
	Bigger $\epsilon$ means more robustness to adversarial examples. 
	The classification accuracies are reported in the last row as reference. 
	We use our implementation of PointNet in PyTorch~\cite{PyTorch} for generating adversarial examples.}
\label{tab:epsilon}
\end{table}
\vspace{-0.6in}


\subsection{Comparison to Prior Work}
\label{sec:comparison}
We compare our MVCNN result with prior works in Table~\ref{tab:compare_acc}. 
The results are grouped by the input type. 
For multi-view image-based models, our MVCNN achieves 95.0\% per instance accuracy, which is the best result between all competing approaches. 
Rotation Net~\cite{kanezaki2016rotationnet}, which predicts the object pose and class labels at the same time, is 0.2\% worse than our MVCNN. 
Dominant Set Clustering~\cite{wang2017dominant} works by clustering image features across views and pooling within the clusters.
Its performance is 1.0\% lower than RotationNet. 
MVCNN-MultiRes~\cite{qi2016volumetric} is the most related to our work. 
They showed that MVCNN with sphere rendering can achieve better accuracy than voxel-based network, suggesting that there is room for improvement in VoxNet. 
Our VoxNet experiment corroborates to this conclusion. 
Furthermore, MVCNN-MultiRes uses images in multiple resolutions to boost its performance. 

For point-based methods, PointNet~\cite{pointnet} and DeepSets~\cite{zaheer2017deep} use symmetric functions, 
i.e. max/mean pooling layers, to generate permutation invariant point cloud descriptions.
DynamicGraph~\cite{wang2018dynamic} builds upon PointNet by performing symmetric function aggregations on points within a neighborhood, instead of the whole set.
Such neighborhood is computed dynamically by building nearest neighbor graph using distances defined in the feature space. 
Similarly, Kd-Networks~\cite{Klokov_2017_ICCV} work by precomputing a graph induced by a binary spatial partitioning tree and use it to apply local linear operations.
The best point-based method is 2.8\% less accurate then our MVCNN.

For voxel-based methods, VoxNet~\cite{voxnet} and 3DShapeNets~\cite{modelnet} work by applying 3D convolutions on voxels. 
ORION~\cite{sedaghat2016orientation} is based on VoxNet but predicts the orientation in addition to class labels. 
OctNet~\cite{Riegler2017CVPR} and O-CNN~\cite{Wang-2017-OCNN} are able to process higher resolution grids by using an octree representation.
FusionNet~\cite{hegde2016fusionnet} combines the voxel and image representations to improve the performance to 90.8\%. 
Our experiments in Section~\ref{sec:representations} suggests that since MVCNN already has 95.0\% 
accuracy the benefit of combining different representations is not effective.

\begin{table}[h!]
\centering
\footnotesize
\begin{tabular}{c|c|c|c}
Model & Input & Per class acc. & Per ins. acc. \\
\hline
\textbf{Our MVCNN} & \multirow{6}{*}{Images} & \textbf{92.4} & \textbf{95.0} \\
RotationNet~\cite{kanezaki2016rotationnet} &  & - & 94.8 \\
Dominant Set Clustering~\cite{wang2017dominant} &  & - & 93.8 \\
MVCNN-MultiRes~\cite{qi2016volumetric} &  & 91.4 & 93.8 \\
PANORAMA-NN~\cite{sfikas2017exploiting} &  &  & 90.7 \\
MVCNN~\cite{mvcnn} &  & 90.1 & 90.1 \\
\hline
DynamicGraph~\cite{wang2018dynamic} & \multirow{6}{*}{PC} & 90.2 & 92.2 \\
Kd-Networks~\cite{Klokov_2017_ICCV} &  & - & 91.8 \\
LocalFeatureNet~\cite{shen2017neighbors} &  & - & 90.8 \\
PointNet++~\cite{Qi:PointNet++:2017} &  & - & 90.7 \\
DeepSets~\cite{zaheer2017deep} &  & - & 90.0 \\
PointNet~\cite{pointnet} &  & 86.2 & 89.2 \\
\hline
VRN Single~\cite{BrockLRW16} & \multirow{5}{*}{Voxels} & - & 91.3 \\
O-CNN~\cite{Wang-2017-OCNN} &  &  &  90.6\\
ORION~\cite{sedaghat2016orientation} &  & - & 89.7 \\
VoxNet~\cite{voxnet} &  & - & 83.0 \\
3DShapeNets~\cite{modelnet} &  & 77.3 & 84.7 \\
\hline
PointNet++~\cite{Qi:PointNet++:2017} & PC+Normal & - & 91.9 \\
FusionNet~\cite{hegde2016fusionnet} & Voxels+Images & - & 90.8 \\
\end{tabular}
\caption{Accuracy (\%) of state-of-the-art methods with different 3D representations.
	PC refers to point clouds. The order is grouped by input type and sorted by accuracy.} 
%
\vspace{-0.35in}
\label{tab:compare_acc}
\end{table}

\section{Conclusion}
\vspace{-0.2in}
\label{s:conclusion}
We investigated on different representations and models for 3D shape classification task, which resulted in a new state-of-the-art on the ModelNet40 benchmark. We analyzed the generalization of MVCNN, PointNet, and VoxNet by varying the number of training examples. Our results indicate that multiview-based methods provide better generalizability and outperform other methods on the full dataset even without ImageNet pretraining or training with binary silhouettes. We also analyzed cross-modal distillation and showed improvements on VoxNet and PointNet by distilling knowledge from MVCNN. Finally, we analyzed the robustness of the models to adversarial perturbations and concluded that point-based networks are more robust to point perturbations, while multi-view and voxel-based networks can be fooled by imperceptible perturbations.

\paragraph{\textbf{Acknowledgment}}
We acknowledge support from NSF (\#1617917, \#1749833) and the MassTech Collaborative grant for funding the UMass GPU cluster.

\bibliographystyle{splncs04}
\bibliography{mrnet}

\pagebreak
\appendix
\section{A Deeper Look at 3D Shape Classifiers: \\Supplementary Material}
In the supplementary material we include (i) accuracy as a function of the number of views in a MVCNN model, (ii) examples of rendered images from 12 views, (iii) enhanced adversarial examples, and (iv) class confusion matrices of different models.

\subsection{Effect of number of views for MVCNN}
\begin{figure}[h!]
\centering
	\begin{tabular}{ccc}
        \includegraphics[clip, width=0.49\linewidth]{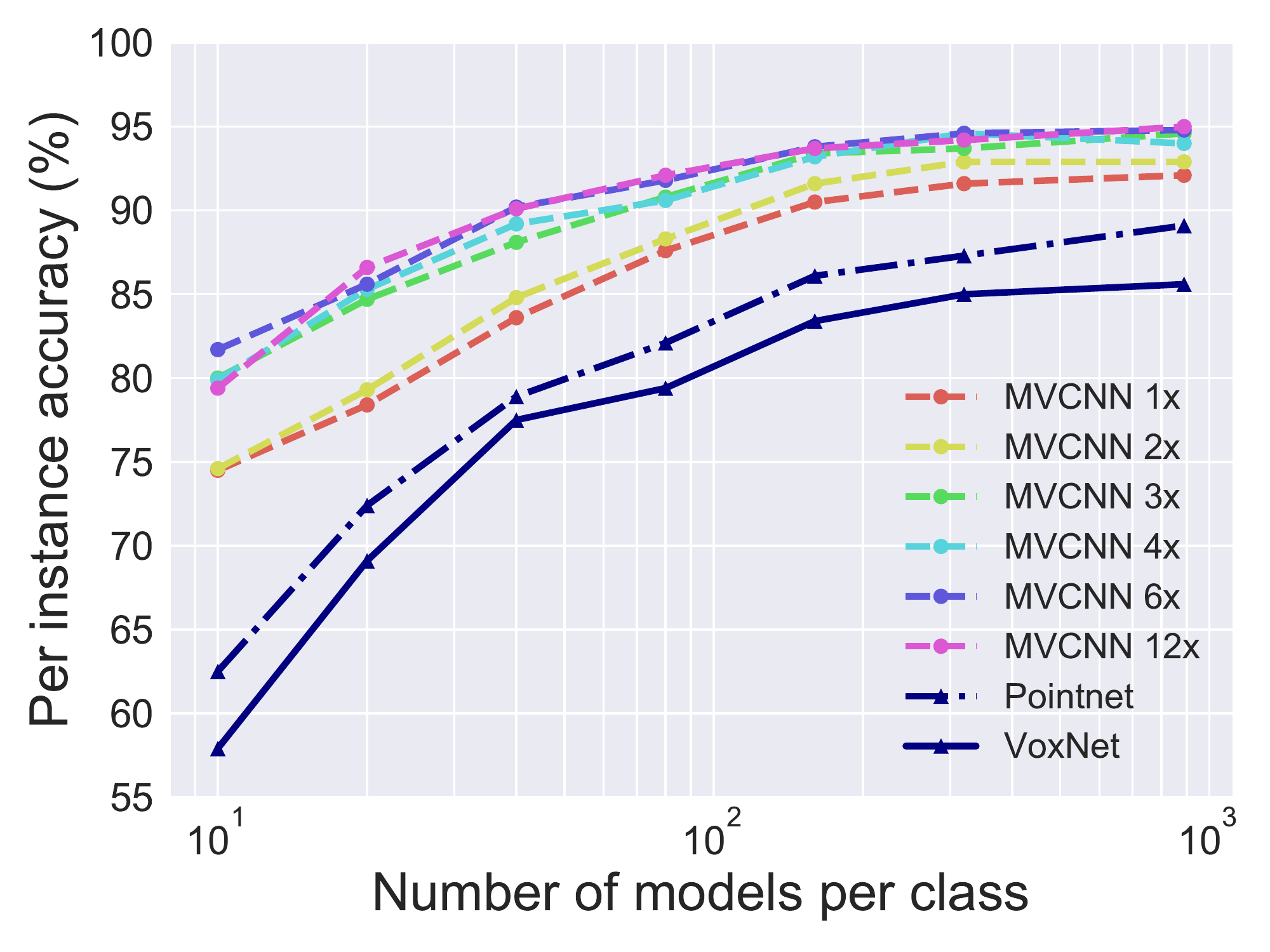}&
        \includegraphics[clip, width=0.49\linewidth]{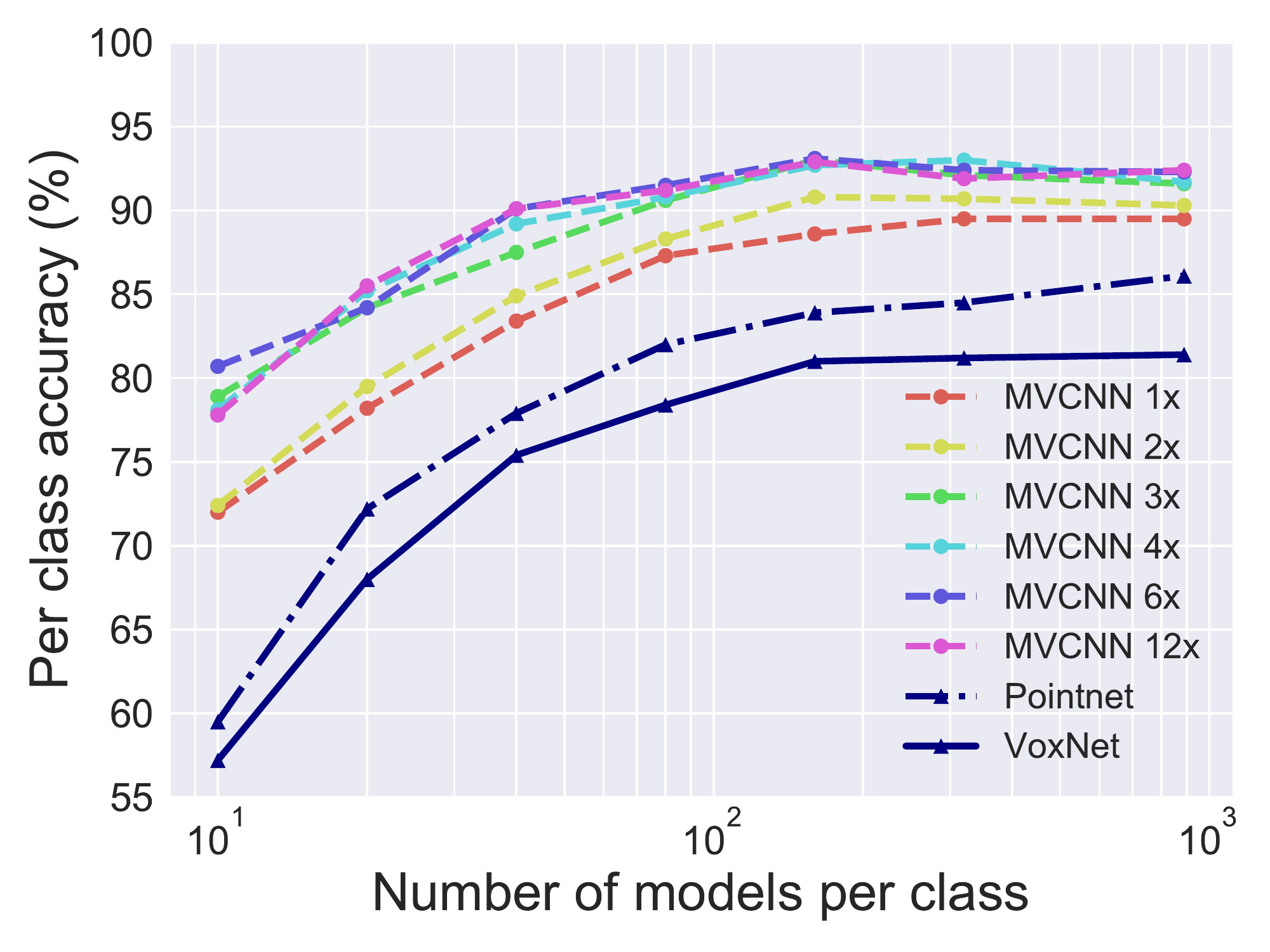}
	\end{tabular}
\caption{Accuracy of the MVCNN with different number of views (1x uses one view, 2x uses two views, etc.)
PointNet and VoxNet results are also included for comparison. MVCNN with one view outperforms both the VoxNet and PointNet.}
\label{fig:views}
\end{figure}

We investigate the effect of the number of views on the accuracy of the model. 
The results using different number of views $k \in \{1,2,3,4,6,12\}$ are shown in Figure~\ref{fig:views}. 
The views are selected radially symmetrically, i.e., the object is successively rotated by $2\pi/k$ for generating one of $k$ views.
Even with one view the MVCNN outperforms both VoxNet and PointNet. 
The performance is close to optimal with only four views per model.

\subsection{Comparison of rendering}
Figure~\ref{fig:shaded} shows the shaded images from 12 views with our rendering (top two rows) and a comparison to the rendering used in~\cite{mvcnn} (bottom two rows). 

\newcommand*{\imw}{0.141}
\begin{figure}[h!]
\centering
	\begin{tabular}{cccccc}
        \includegraphics[clip, width=0.16\linewidth]{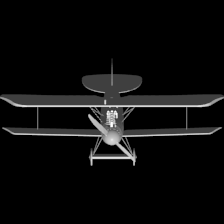}&
        \includegraphics[clip, width=0.16\linewidth]{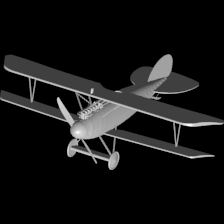}&
        \includegraphics[clip, width=0.16\linewidth]{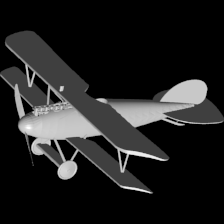}&
        \includegraphics[clip, width=0.16\linewidth]{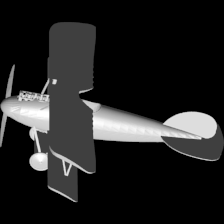}&
        \includegraphics[clip, width=0.16\linewidth]{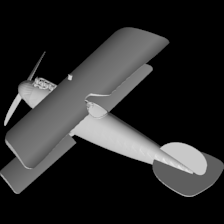}&
        \includegraphics[clip, width=0.16\linewidth]{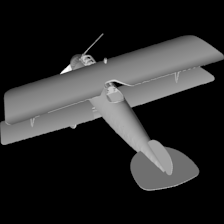}\\
        \includegraphics[clip, width=0.16\linewidth]{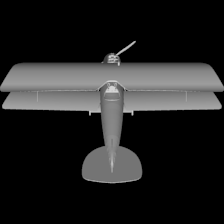}&
        \includegraphics[clip, width=0.16\linewidth]{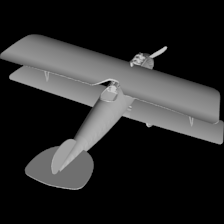}&
        \includegraphics[clip, width=0.16\linewidth]{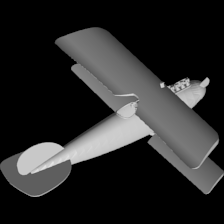}&
        \includegraphics[clip, width=0.16\linewidth]{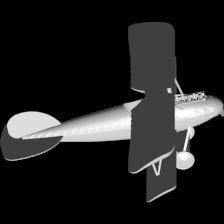}&
        \includegraphics[clip, width=0.16\linewidth]{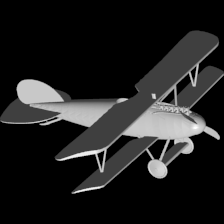}&
        \includegraphics[clip, width=0.16\linewidth]{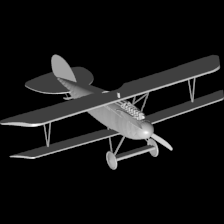}\\

        \fbox{%
        	\includegraphics[clip, width=\imw\linewidth]{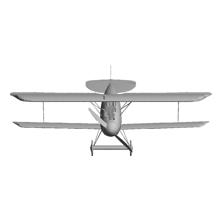}%
			}&
        \fbox{%
        	\includegraphics[clip, width=\imw\linewidth]{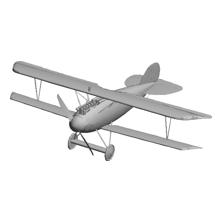}%
			}&
        \fbox{%
        \includegraphics[clip, width=\imw\linewidth]{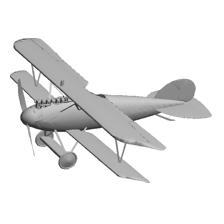}%
        	}&
        \fbox{%
        \includegraphics[clip, width=\imw\linewidth]{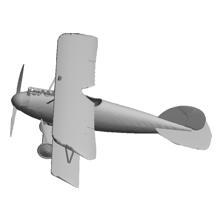}%
        	}&
        \fbox{%
        \includegraphics[clip, width=\imw\linewidth]{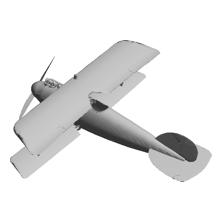}%
        	}&
        \fbox{%
        \includegraphics[clip, width=\imw\linewidth]{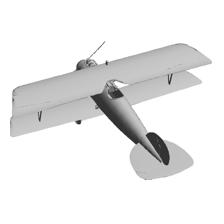}%
        	}\\
        \fbox{%
        \includegraphics[clip, width=\imw\linewidth]{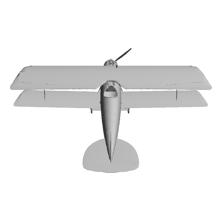}%
        	}&
        \fbox{%
        \includegraphics[clip, width=\imw\linewidth]{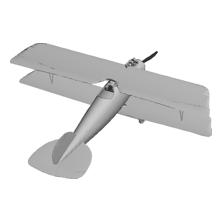}%
        	}&
        \fbox{%
        \includegraphics[clip, width=\imw\linewidth]{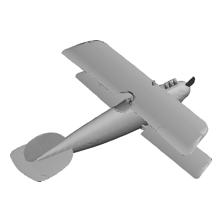}%
        	}&
        \fbox{%
        \includegraphics[clip, width=\imw\linewidth]{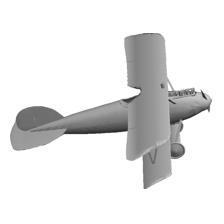}%
        	}&
        \fbox{%
        \includegraphics[clip, width=\imw\linewidth]{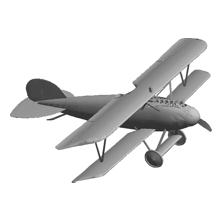}%
        	}&
        \fbox{%
        \includegraphics[clip, width=\imw\linewidth]{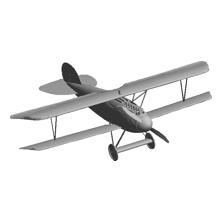}%
        	}\\
	\end{tabular}
\caption{Shaded images from 12 views our rendering (top two rows) and those from MVCNN~\cite{mvcnn} rendering (bottom two rows). The images from left to right and top to bottom the images are ordered according to the orientation in increments of 30 degrees.}
\label{fig:shaded}
\end{figure}

\subsection{Enhanced adversarial examples}
\begin{figure}[h!]
\centering
	\begin{tabular}{cccc}
	         \includegraphics[clip, width=0.20\linewidth]{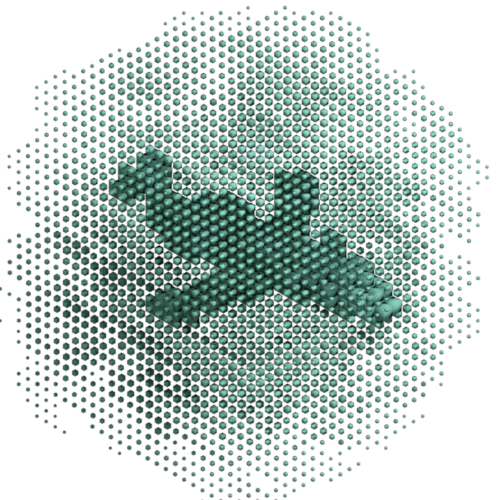}&
        \includegraphics[clip, width=0.20\linewidth]{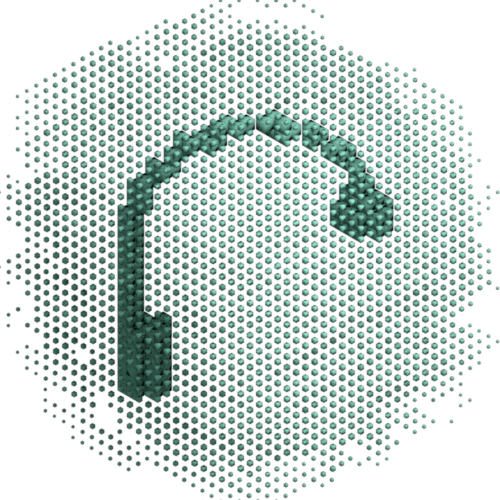}&
        \includegraphics[clip, width=0.20\linewidth]{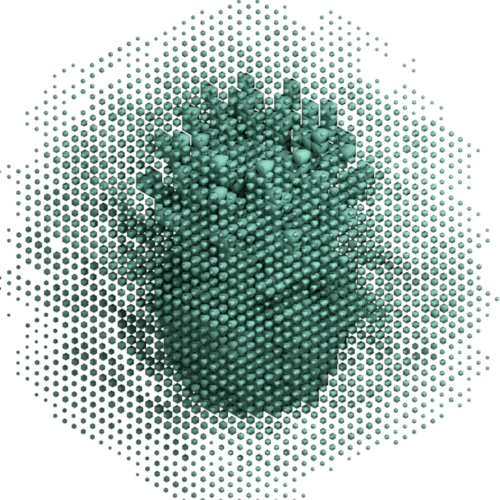}&
        \includegraphics[clip, width=0.20\linewidth]{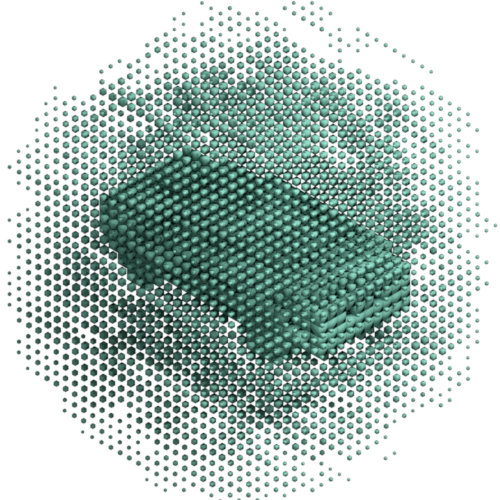}\\
         $\epsilon=0.07$ &  $\epsilon=0.02$ &  $\epsilon=0.2$ &  $\epsilon=0.1$ \\
        \includegraphics[clip, width=0.20\linewidth]{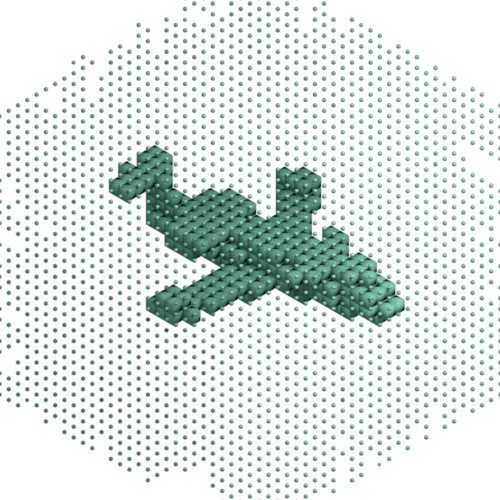}&
        \includegraphics[clip, width=0.20\linewidth]{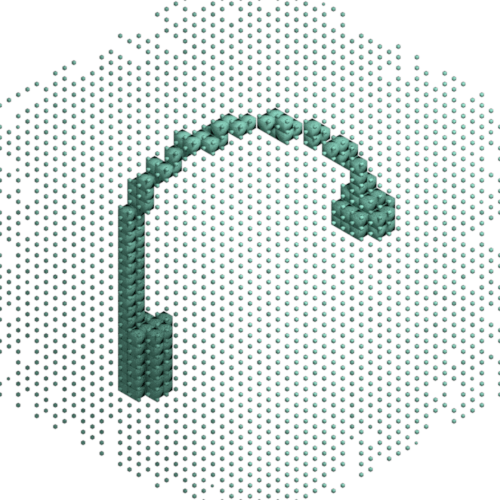}&
        \includegraphics[clip, width=0.20\linewidth]{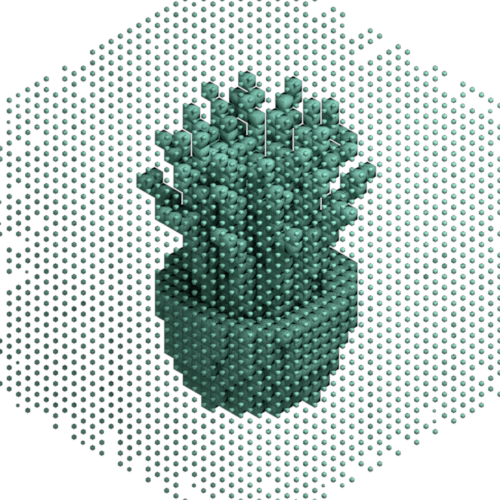}&
        \includegraphics[clip, width=0.20\linewidth]{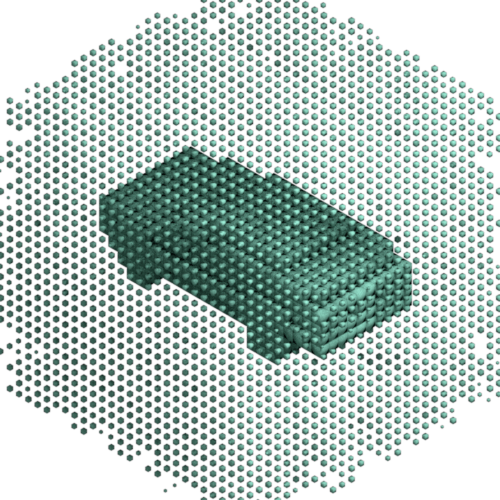}\\
         $\epsilon=0.003$ &  $\epsilon=0.004$ &  $\epsilon=0.007$ &  $\epsilon=0.02$ 
	\end{tabular}
\caption{Adversarial examples of VoxNet and VoxMVCNN with the adversarial noise enhanced for visibility. Voxel size represents occupancy level raised to the power of $\frac{1}{3}$. Original $\epsilon$ are shown below each image. \textbf{First row:} VoxNet, \textbf{Second row:} VoxMVCNN.}
\label{fig:gallery_enhanced}
\end{figure}

Figure~\ref{fig:gallery_enhanced} shows the adversarial examples of VoxNet and VoxMVCNN with the adversarial noise enhanced for improved visibility. 
Since the $\epsilon$ are small for adversarial examples, we raise the noise to $\epsilon^{1/3}$. 
These two rows correspond to the last two columns in Figure~\ref{fig:gallery} in the paper.

\subsection{Confusion matrices}
\begin{figure}[h!]
\centering
        \includegraphics[clip, width=0.60\linewidth]{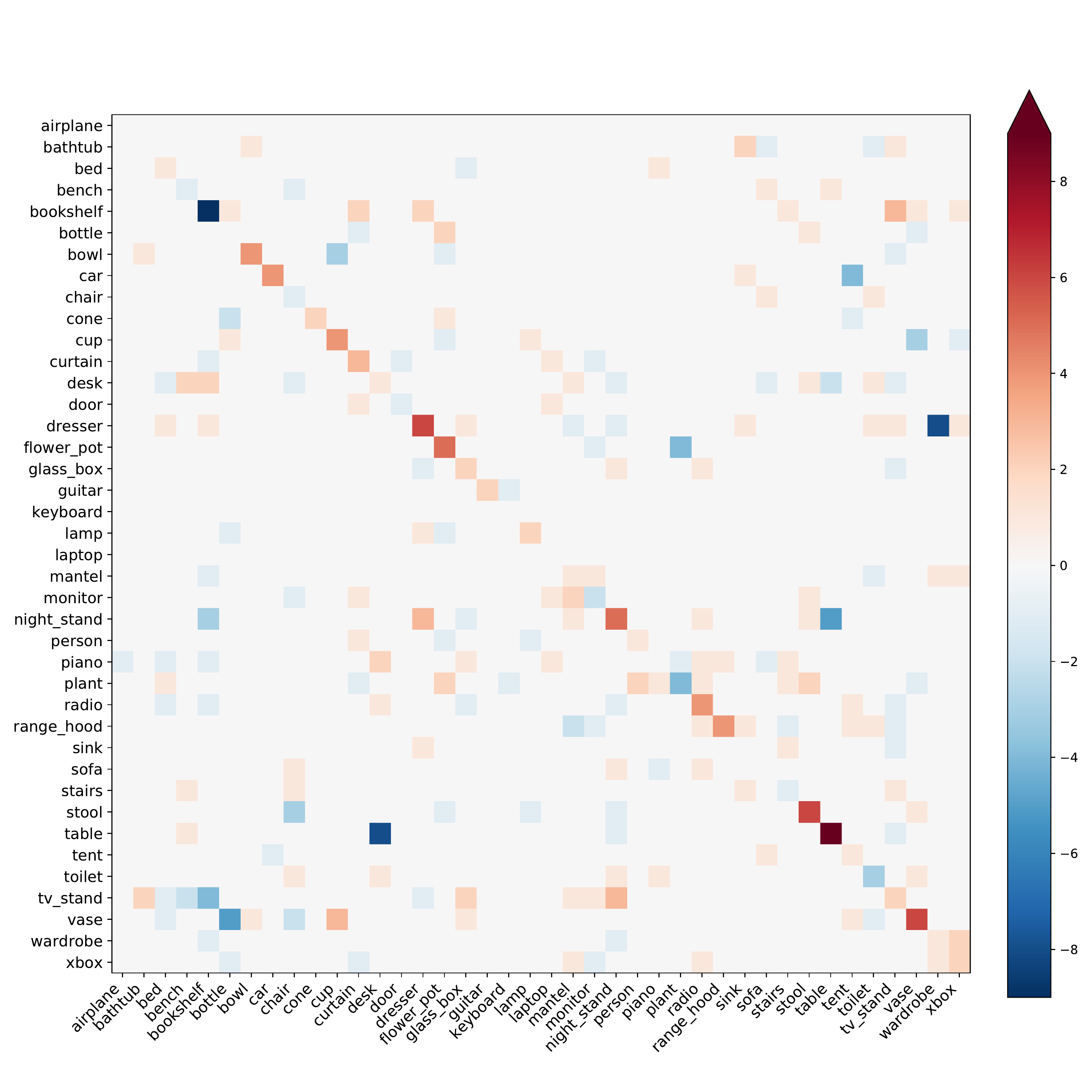}\\
        (c) PointNet - VoxNet
\caption{The differences of confusion matrices of different models. Red represents positive values, blue represents negative values, while white means zero values as seen in the color bar. The units here are the number of test models. In panel (a) and (b) most values are positive in main diagonals and zero or negative in off-diagonals showing that MVCNN outperforms other methods. Panel (c) shows that PointNet and VoxNet make complementary mistakes.}
\label{fig:confusion}
\end{figure}

Here we compare the class confusion matrices of different models.
Figure~\ref{fig:confusion} shows the differences of the confusion matrices between (a) MVCNN and PointNet, (b) MVCNN and VoxNet, and (c) PointNet and VoxNet. The difference of confusion matrices (A-B) is obtained by subtracting the confusion matrix of method B from method A. Higher values in main diagonals and lower values in off-diagonals indicate method A has better performance than method B.
In Figure~\ref{fig:confusion}(a-b) we show that MVCNN outperforms PointNet and VoxNet.
There are only a few confusions where MVCNN is worse then VoxNet or PointNet, e.g. $flower\ pot \rightarrow plant$. These classes are hard to be distinguished even for humans. Figure~\ref{fig:confusion}(c) shows PointNet has better performance than VoxNet in some cases, e.g. $dresser \rightarrow wardrobe$, but also has many confusions in other cases. 

\begin{figure}[h!]
\centering
        \includegraphics[clip, width=0.60\linewidth]{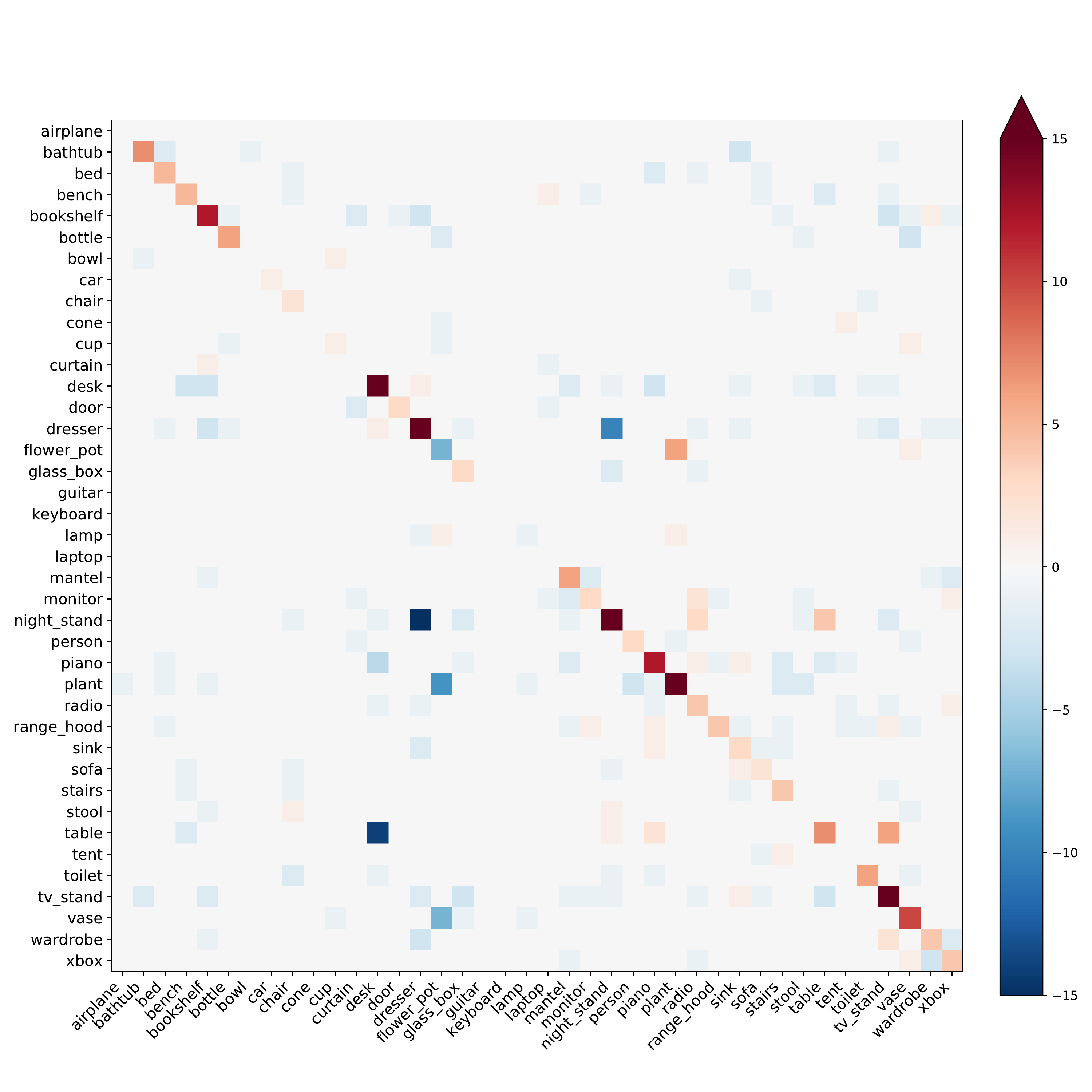}\\
        (a) Differnce MVCNN - PointNet  \\
\end{figure}

\begin{figure}[h!]
\centering
        \includegraphics[clip, width=0.60\linewidth]{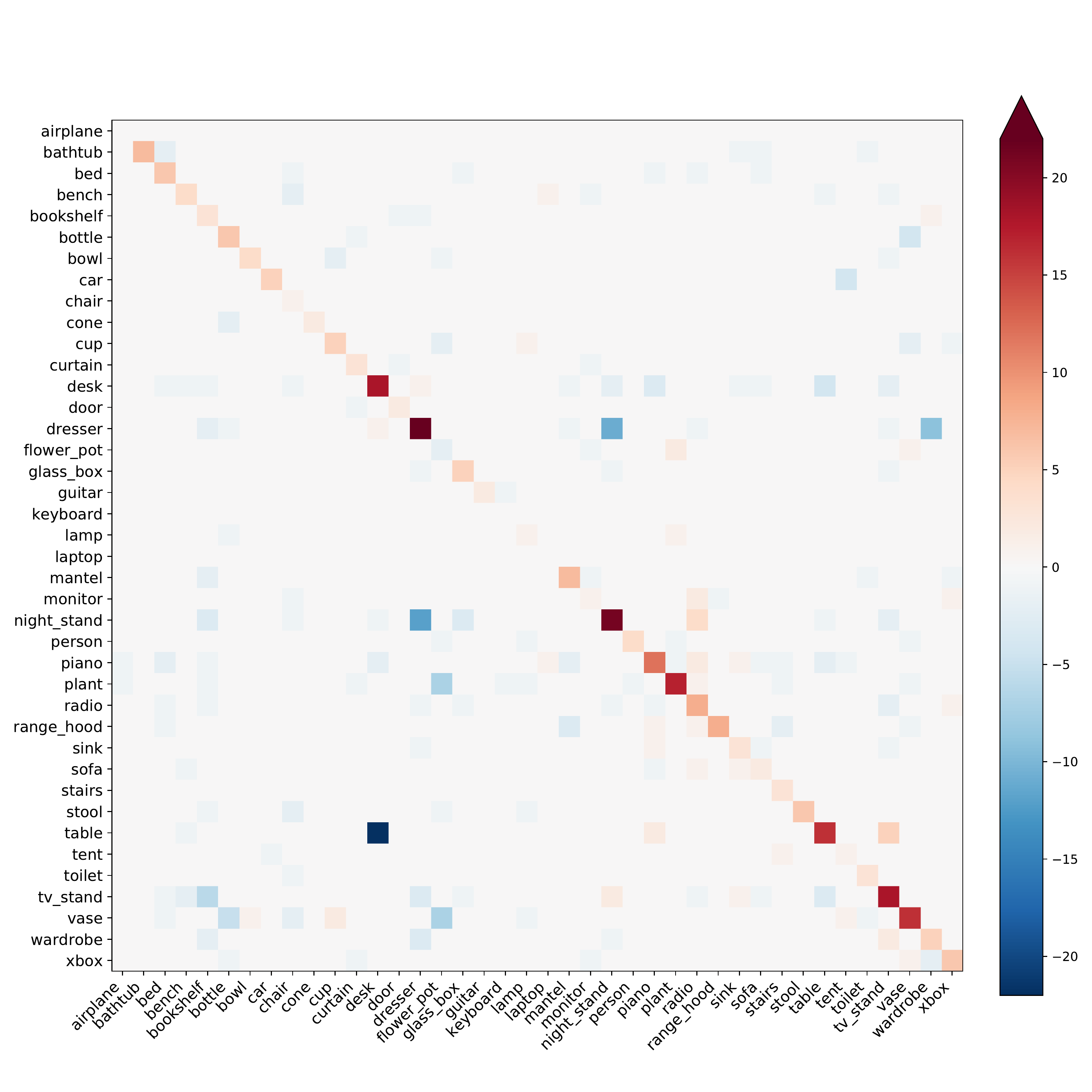}\\
        (b) MVCNN - VoxNet \\
\end{figure}

\end{document}